\documentclass[11pt]{article}
\usepackage{dblfloatfix}
\usepackage{acl}

\usepackage{times}
\usepackage{latexsym}
\usepackage[table]{xcolor}
\usepackage{subcaption}
\usepackage[T1]{fontenc}

\usepackage[utf8]{inputenc}

\usepackage{microtype}
\usepackage{multirow} 
\usepackage{booktabs}
\usepackage{amssymb}
\usepackage{graphicx}
\usepackage[most]{tcolorbox}

\usepackage{inconsolata}

\usepackage{graphicx}
\usepackage{amsmath}
\usepackage{xspace}
\newcommand{\system}[1]{\textsc{#1}\xspace} 
\newcommand{\ourmethod}{\system{CoTEvol}}

\title{\ourmethod: Self-Evolving Chain-of-Thoughts for Data Synthesis in Mathematical Reasoning}

\newcommand{\aspace}{\hspace{0.5em}}
\newcommand{\fudan}{$^{1}$}
\newcommand{\sii}{$^{2}$}
\newcommand{\hk}{$^{3}$}
\newcommand{\monash}{$^{4}$}
\newcommand{\research}{$^{5}$}
\newcommand{\szy}{$^{6}$}

\author{
Zhuo Wang\fudan\sii\thanks{These authors contributed equally.} \aspace
Zhuo Zhang\research\footnotemark[1] \aspace
Yafu Li\hk \aspace
Yu Cheng\sii\hk\thanks{Corresponding authors.} \aspace
Lizhen Qu\monash\footnotemark[2] \aspace
Zenglin Xu\fudan\szy\footnotemark[2]\\
\fudan{}Fudan University \quad \sii{}Shanghai Innovation Institute \quad
\hk{}The Chinese University of Hong Kong \\
\monash{}Monash University \quad \research{}Independent Researcher \quad \szy{}Shanghai Academy of AI for Science \\
\texttt{zwang24}@m.fudan.edu.cn \quad \texttt{\{iezhuo17, yafuly\}@gmail.com} \\
\texttt{chengyu}@cse.cuhk.edu.hk \quad \texttt{Lizhen.Qu}@monash.edu \quad \texttt{zenglinxu}@fudan.edu.cn
}

\begin{document}
\maketitle

\begin{abstract}
Large Language Models (LLMs) exhibit strong mathematical reasoning when trained on high-quality Chain-of-Thought (CoT) that articulates intermediate steps, yet costly CoT curation hinders further progress.
While existing remedies such as distillation from stronger LLMs and self-synthesis based on test-time search alleviate this issue, they often suffer from diminishing returns or high computing overhead.
In this work, we propose \ourmethod, a genetic evolutionary framework that casts CoT generation as a population-based search over reasoning trajectories. 
Candidate trajectories are iteratively evolved through reflective global crossover at the trajectory level and local mutation guided by uncertainty at the step level, enabling holistic recombination and fine-grained refinement.
Lightweight, task-aware fitness functions are designed to guide the evolutionary process toward accurate and diverse reasoning. 
Empirically, \ourmethod improves correct-CoT synthesis success by over 30\% and enhances structural diversity, with markedly improved efficiency. LLMs trained on these evolutionary CoT data achieve an average gain of 6.6\% across eight math benchmarks, outperforming previous distillation and self-synthesis approaches. 
These results underscore the promise of evolutionary CoT synthesis as a scalable and effective method for mathematical reasoning tasks. 
\end{abstract}


\section{Introduction}
Large Language Models (LLMs) have exhibited remarkable reasoning capabilities in solving complex mathematical tasks when prompted to generate step-by-step solutions~\cite{wei2022chain,renze2024self,jaech2024openai,deepseek-r1}.
One of the key reasons is the quality and availability of curated Chain-of-Thought (CoT) training data~\cite{luo2024improve}, which empowers LLMs to construct logically coherent reasoning trajectories.
However, such data are scarce and costly to produce due to the labor-intensive and time-consuming annotation process, which limits the advancement of reasoning-oriented LLMs.

\begin{figure}[t]
  \centering
  \includegraphics[width=0.4\textwidth]{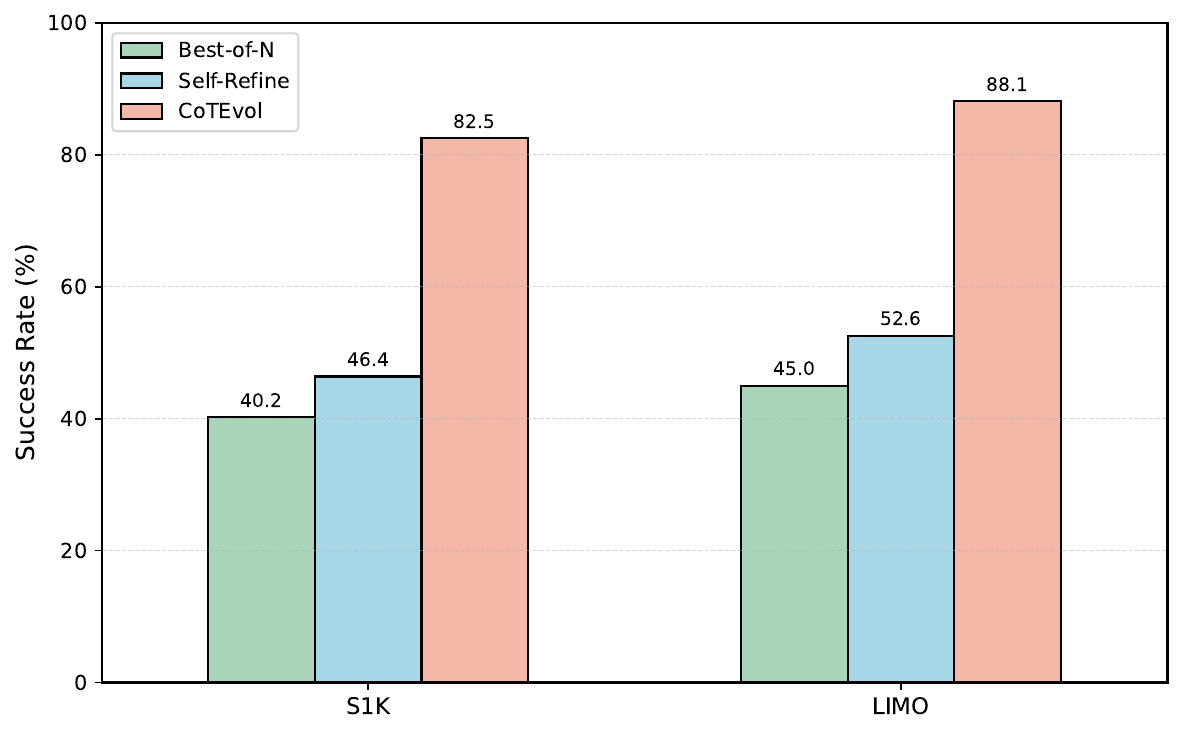}  
  \caption{Comparison of CoT synthesis success rates of Best-of-N, Self-Refine, and \ourmethod on the S1K and LIMO training datasets.}
  \vspace{-0.5cm}
  \label{fig:small_pic}
\end{figure}

Existing studies address data scarcity in two main directions. One distills reasoning traces from frontier LLMs, such as DeepSeek-R1~\cite{deepseek-r1} or OpenAI o1~\cite{jaech2024openai}, while the other relies on test-time scaling to identify effective reasoning trajectories through allocating more computational resources~\cite{snell2024scaling,levi2024simple,best_of_n}. Although both approaches partially alleviate this problem, they suffer from inherent limitations: distillation exhibits diminishing returns\footnote{For example, OpenMathInstruct-2~\cite{toshniwal2024openmathinstruct} reports only a 3.9\% improvement on MATH despite using 8$\times$ more data.} and may propagate biases from teacher models~\cite{guo2024bias}, while test-time scaling relies on extensive repeated sampling~\cite{diao2023active,madaan2023self}, with largely independent trials that hinder the reuse of intermediate reasoning signals and often cause the search to stagnate in low-quality regions, resulting in poor computational efficiency.

To overcome these limitations, we reformulate CoT synthesis as a heuristic search and combinatorial optimization problem. Given a question, it aims to efficiently discover CoTs as reasoning trajectories that lead to the correct answer. This optimization perspective naturally motivates the use of genetic algorithms (GAs), whose crossover and mutation operators can be explicitly adapted to explore complex solution spaces. The resulting trajectories can then be used to fine-tune LLMs, improving their reasoning capabilities.

Inspired by this, we propose \ourmethod, a \textit{novel} self-evolving CoT synthesis framework that adapt GAs to work directly on LLM-generated token sequences, overcoming their reliance on discrete symbolic representations. 
\ourmethod treats reasoning trajectories as individuals in a population and iteratively improves them through genetic operations. Specifically, we introduce reflective global crossover, which constructs high-level feedback to guide the recombination of complementary reasoning structures, and uncertainty-guided local mutation, which identifies unstable reasoning steps via entropy and refines ambiguous parts of the reasoning process. Lightweight, task-aware fitness functions further guide evolution by jointly evaluating solution correctness and reasoning structure. Through this population-based evolution, \ourmethod efficiently discovers a large set of CoTs that yield correct answers on mathematical reasoning tasks. As shown in Figure~\ref{fig:small_pic}, our method improves solution correctness in the synthesized training data by over \textit{30\%} while requiring less computational cost. Fine-tuning models on these evolved CoTs yields an average \textit{6.6\%} gain in inference accuracy. Overall, \ourmethod offers a cost-effective way for mathematical reasoning. Our main contributions can be summarized as follows:

\begin{itemize}

    \item We propose \ourmethod, a self-evolutionary CoT synthesis framework inspired by genetic algorithms, which exhaustively exploits the intrinsic reasoning potential of LLMs.
    \item We introduce reflective global crossover for structural recombination and uncertainty-guided local mutation for targeted refinement, both navigated by lightweight, task-aware fitness functions.
    \item Experiments show that \ourmethod improves synthesis correctness by over 30\% and yields a 6.6\% average accuracy gain across eight benchmarks, demonstrating an effective and cost-efficient approach to CoT synthesis. 

    


\end{itemize}

\section{Related Work}
\paragraph{Mathematical Data Synthesis.}
Synthetic math data generation~\cite{madaan2023self,rstar,deepseek-r1,deepmath103k} is an effective way to improve mathematical reasoning when human annotations are scarce~\cite{villalobos2022will}. Existing methods fall into two categories: (1) \emph{distillation} from powerful LLMs, producing high-quality datasets (e.g., OpenMathInstruct-2~\cite{toshniwal2024openmathinstruct}) but incurring high computational cost and potential bias~\cite{rstar,guo2024bias}; and (2) \emph{self-synthesis} via test-time search, generating CoTs without external supervision. Representative methods include Best-of-N~\cite{best_of_n,diao2023active}, which samples multiple trajectories and selects the best via an external reward model, and Self-Refine~\cite{madaan2023self}, which iteratively improves CoTs based on model feedback. 

However, these approaches often rely on external reward models and are constrained by the capabilities of the policy LLM, leading to low sampling efficiency and limited gains. In contrast, we propose \ourmethod to elicit the model’s intrinsic reasoning potential while selectively invoking stronger LLMs only for hard cases, thereby resulting in a practical and scalable CoT synthesis framework with controllable computational cost.

\paragraph{Language-based Genetic Algorithms for Mathematical Reasoning.}
Genetic algorithms (GAs)~\cite{mitchell1998introduction} have demonstrated effectiveness in exploring complex solution spaces under diverse constraints~\cite{yao2025multi}. With LLMs, GAs can operate in a language-centric paradigm, representing individuals as natural language and performing crossover and mutation through generation. Prior work has applied these ideas to inference-time optimization. FunSearch~\cite{fawzi2023funsearch} evolves code-based programs to discover algorithmic solutions. Mind Evolution~\cite{lee2025evolving} applies GAs framework to planning tasks  such as TravelPlanner, and Meeting Planning. 

By contrast, our approach evolves textual CoT trajectories for training data synthesis rather than inference-time optimization. Furthermore, We design biologically inspired and fine-grained crossover and mutation operators, where crossover performs global recombination and mutation introduces targeted local variations, distinguishing our method from prior work in math reasoning.

\begin{figure*}[!t]
  \centering  
  \includegraphics[width=1.0\textwidth]{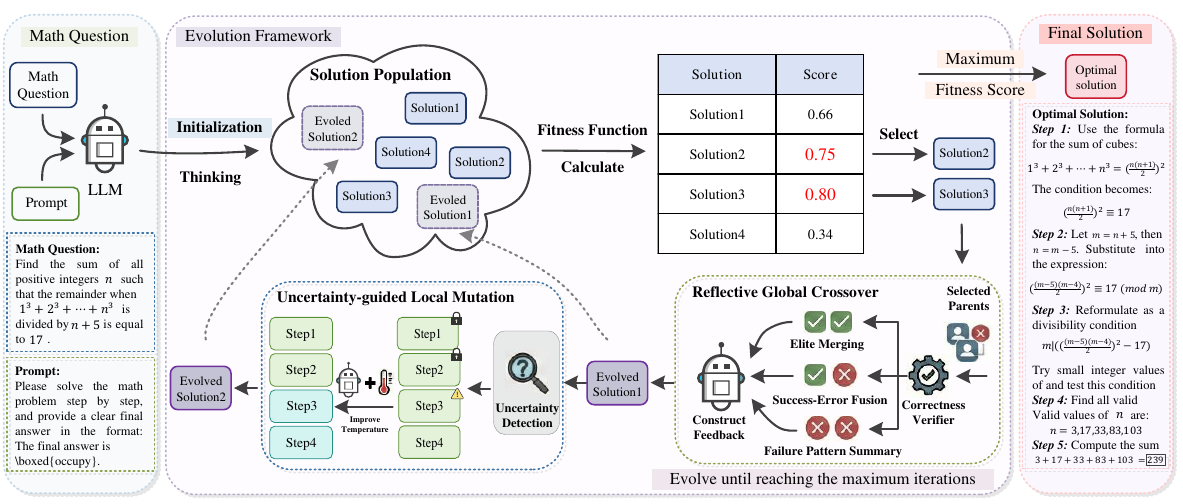}
  \vspace{-14pt}
  \caption{The framework of \ourmethod. \ourmethod evolves mathematical chain-of-thoughts through iterative fitness-based selection, reflective global crossover, and uncertainty-guided local mutation.}
  \vspace{-0.4cm}
  \label{fig:overview}
\end{figure*}

\section{Methodology}

This section presents \ourmethod, a novel self-supervised CoT data synthesis framework that leverages fine-grained genetic operators and the self-reflective capabilities of LLMs to unlock the potential of test-time computation. Subsequently, these self-evolved CoT data are employed to train LLMs using supervised fine-tuning, enhancing performance on mathematical reasoning tasks.

\subsection{Framework}
\ourmethod has three components: population management, an adaptive evolution strategy, and a fitness function. Figure \ref{fig:overview} presents the overall framework, which evolves reasoning trajectories through a population-based optimization process. The framework begins with population initialization. In each iteration, high-fitness solutions are selected as parents and refined through the strategy at two complementary levels: (i) \emph{reflective global crossover}, which performs global knowledge transfer across reasoning paths via reflective feedback, and (ii) \emph{uncertainty-guided local mutation}, which conducts targeted local refinement by revising high-uncertainty reasoning steps. The resulting offspring are incorporated into the population while maintaining a fixed population size. Iterations continue until convergence or a maximum number of steps, after which the highest-fitness solution is selected as the CoT data for supervised fine-tuning.

\subsection{Fitness Function}
The fitness function guides the evolution process by providing scalar feedback that evaluates the quality of solutions. 
Recent advances in reinforcement learning from verifiable rewards (RLVR) \cite{deepseek-r1,simplerl} have highlighted the potential of structured reward functions in enhancing the reasoning capabilities of LLMs. 
Inspired by prior work~\cite{luong2024reft}, we design three rule-based verifiers tailored to evaluate mathematical CoT data.

\paragraph{Answer Correctness Verifier.} This verifier assesses whether a CoT solution arrives at the correct answer for a given problem. A correct solution receives a fitness score $\mathcal{R}_{ac}$ of 1, while an incorrect one receives 0. To improve robustness in borderline cases, we adopt a strategy from prior work~\cite{luong2024reft} by assigning a score of 0.5 to answers that are numerically interpretable but incorrect.

\paragraph{Format Matching Verifier.} This verifier checks whether the generated answer adheres to a predefined format. Specifically, our method requires all final answers to be enclosed in \texttt{/boxed$\{\}$} notation. If the format is correctly parsed and satisfies the requirement, the reward $\mathcal{R}_{fmt}$ of 0.5 is given; otherwise, the reward $\mathcal{R}_{fmt}$ is set to 0.

\paragraph{Length-based Reward Verifier.} This verifier adjusts rewards based on the length of the generated solution, aiming to balance efficiency with completeness. Concise and informative reasoning is encouraged, while unnecessarily verbose outputs are mildly penalized. Following prior work~\cite{openr1}, we adopt a cosine-scaling reward function that differentiates between correct and incorrect answers. For a candidate solution $(q_i,x_i, \hat{y_i}, y_i)$ within a population $P$, the reward is computed as:
{
\begin{equation}
\label{eq:R_len} 
\resizebox{\linewidth}{!}{
    $\mathcal{R}_{\text{len}} =
    \begin{cases}
    \text{C}_{\min} + 0.5 \cdot (\text{C}_{\max} - \text{C}_{\min}) \cdot (1 + \cos(\pi \cdot \frac{L}{L_{\max}})) & \text{if } \hat{y_i} = y_i \\
    \text{W}_{\min} + 0.5 \cdot (\text{W}_{\max} - \text{W}_{\min}) \cdot (1 + \cos(\pi \cdot \frac{L}{L_{\max}})) & \text{if } \hat{y_i} \neq y_i
    \end{cases}$
}
\end{equation}
} where $q_i$ is the $i$-th question, $L$ denotes the length of the current CoT solution $x_i$, $\hat{y_i}$ denotes the predicted answer, $y_i$ is ground truth, and $L_{\max}$ is the maximum content length within the current population. The terms $\text{C}_{\min}$ and $\text{C}_{\max}$ define the reward range for correct answers, while $\text{W}_{\min}$ and $\text{W}_{\max}$ represent the corresponding bounds for incorrect answers. The cosine factor $\cos\left(\pi \cdot \frac{L}{L_{\max}}\right)$ is introduced to smoothly modulate the effect of length. It maintains progressive penalization while mitigating an excessive preference for shorter outputs.

The final fitness score for candidate individuals is the sum of the scores of the above three verifiers.
\begin{equation}
    \mathcal{R} = \mathcal{R}_{ac} + \mathcal{R}_{fmt} + \mathcal{R}_{len}
\end{equation}



\subsection{Population Management}
Population management serves three core functions in our \ourmethod: (1) Population Initialization: constructing initial candidate solutions before evolution; (2) Elite Selection: selecting individuals based on fitness scores to serve as parents to generate offspring; (3) Population Updates: dynamically controlling population size to ensure effective selection pressure and maintain diversity.

\paragraph{Population Initialization.} 
Given a specific mathematical problem, we prompt LLMs to independently generate $N_{pop}$ initial individuals. To balance solution quality and diversity, we set the sampling temperature to 0.6, encouraging variation among generated responses.
After generation, we perform post-processing to filter out redundant or low-quality samples. Specifically, for any pair of individuals with a ROUGE-L~\cite{rougeL} score greater than 0.7, we randomly retain one to eliminate semantic duplicates. Meanwhile, responses that are incomplete, malformed, or indicative of model failure are discarded directly\footnote{For example, if the model crashes or fails to produce a complete, coherent solution.}. This process is repeated until $N_{pop}$ valid and diverse individuals are obtained for initialization.

\paragraph{Elite Selection.} 
To select parent solutions for offspring generation, we adopt the Boltzmann Tournament Selection~\cite{rougeL} strategy. Concretely, we normalize the fitness scores of all individuals using a softmax function to obtain a probability distribution, from which we randomly sample a set of parents. At each generation $t$, we select the top $k$ candidates from the current population $\mathcal{P}_t$ based on their fitness scores. This selection mechanism strikes a balance between exploitation and exploration: high-fitness individuals are more likely to be propagated, improving overall solution quality. Meanwhile, low-fitness individuals retain a non-negligible probability of selection, helping maintain population diversity and mitigate premature convergence.

\paragraph{Population Updates.}
After each evolutionary iteration, we apply fitness-based population control to maintain a fixed population size and keep selection pressure throughout the evolutionary process. Specifically, all individuals are ranked based on their fitness scores, and a selection mechanism is employed to preferentially retain high-quality individuals while discarding less competitive ones.

\subsection{Adaptive Evolution Strategy}\label{sec_cromut}
We propose a novel adaptive evolution strategy for reasoning self-improvement, which combines reflective global crossover for global knowledge transfer with uncertainty-guided local mutation for targeted local exploration. 
By integrating structured feedback and uncertainty-aware refinement, the strategy enables progressive improvement of chain-of-thought data quality.

\paragraph{Reflective Global Crossover.} Given candidate CoTs, parent solutions are first selected according to their fitness scores, which reflect overall solution quality. 
An answer correctness verifier $\mathcal R_{\text{ac}}$ assigns each selected parent a binary correctness label, where $\mathcal R_{\text{ac}}(x) \ge 0.5$ indicates a correct final answer, and $\mathcal R_{\text{ac}}(x) = 0$ otherwise. 
These labels are then used to condition reflection-based feedback during crossover. 

We construct global feedback under three distinct cases: (i) \textbf{Elite Merging:} when both parents are correct, the LLM synthesizes their shared strengths and unique reasoning techniques into a set of \emph{best practices}, guiding the offspring to generate more concise and efficient CoTs; 
(ii) \textbf{Success-Error Fusion:} when one parent is correct and the other incorrect, the LLM extracts key successful strategies from the correct parent while precisely analyzing error patterns in the incorrect parent, reinforcing valid reasoning steps and avoiding known mistakes; 
(iii) \textbf{Failure Pattern Summary:} when both parents are incorrect, the LLM identifies common errors and reasoning pitfalls, producing a \emph{negative knowledge list} that directs the model to avoid these mistakes and explore alternative reasoning paths. The resulting feedback is provided as an additional Information to the LLM, which leverages it to perform crossover and generate the offspring CoT. The corresponding prompts are provided in the Appendix.

\paragraph{Uncertainty-guided Local Mutation.}
To enable fine-grained local exploration within a single CoT, we introduce an uncertainty-driven mutation mechanism that adaptively perturbs unstable reasoning steps. 
The core intuition is that high predictive uncertainty reflects unstable decision points, where alternative reasoning trajectories are more likely to yield improvements.

We first estimate uncertainty at the token level. 
Given the model’s conditional distribution over the vocabulary $\mathcal V$ at token position $j$, the token-level entropy is defined as:
\begin{equation}
H_j = - \sum_{v \in \mathcal V} p(v \mid x_{<j}) \log p(v \mid x_{<j})
\end{equation}
which measures the uncertainty of the next-token prediction conditioned on the preceding context.

A complete solution is represented as an sequence of reasoning steps $x = [x_1, \ldots, x_s]$, where each step $x_s$ spans a contiguous segment of tokens indexed by $T_s$. Token-level uncertainty is then aggregated into step-level entropy by averaging over tokens within the same step:

\begin{equation}
H_s^{\text{step}} = \frac{1}{|T_s|} \sum_{j \in T_s} H_j
\end{equation}
the most unstable reasoning step is identified as:
\begin{equation}
s^{*} = \arg\max_s H_s^{\text{step}}
\end{equation}
starting from the identified step $s^{*}$, we apply mutation by increasing the temperature to encourage broader exploration, while preserving the prefix preceding $s^{*}$. 
Specifically, the mutation temperature is adaptively scaled based on the step entropy:
\begin{equation}
\tau_{\text{mut}} = \tau_0 \, (1 + \lambda H_{s^{*}}^{\text{step}})
\end{equation}
where $\tau_0$ is the base temperature and $\lambda$ controls the strength of entropy-based amplification.

By localizing exploration to high-uncertainty steps, entropy-driven mutation promotes effective refinement of reasoning trajectories without disrupting well-established intermediate conclusions. 

\paragraph{Evolution with no Ground Truth.} 
We define a self-supervised variant of CoTEvol where ground-truth (GT) answers are not available. In this setting, the answer correctness verifier is removed; instead, the LLM performs self-evaluation to score candidate reasoning trajectories, while other verifiers are retained. The crossover operator is guided by self-reflection over parent trajectories, while the mutation operator remains unchanged except that all answer-related content is removed from the prompt. This design enables CoTEvol to operate in a fully self-supervised manner while unlocking the model’s intrinsic reasoning potential.

\section{Experiment}

\subsection{Experiment Setup}

\begin{table*}[]
\centering
\caption{Performance comparison on mathematical benchmarks of models trained separately on the S1K and LIMO datasets using different data synthesis methods, where the best results are highlighted in bold.}\label{tab:main_result}
\vspace{-0.1cm}
\renewcommand{\arraystretch}{1.35}
\resizebox{\linewidth}{!}{
\begin{tabular}{llccccccccc}
\toprule
& &  \textbf{GSM8K} & \textbf{MATH500} & \multirow{2}{*}[1.6ex]{\shortstack[c]{\textbf{Minerva} \\ \textbf{Math}}} & \textbf{GaokaoEn23} & \multirow{2}{*}[1.6ex]{\shortstack[c]{\textbf{Olympiad} \\ \textbf{Bench}}} & \multirow{2}{*}[1.6ex]{\shortstack[c]{\textbf{College} \\ \textbf{Math}}} & \textbf{AIME24} & \textbf{AMC23} & \textbf{AVG.} \\ \midrule

Base Model              & \texttt{Qwen2.5-7b-Instruct}       & 79.6 & 69.4 & 35.7 & 55.3 & 34.7 & 35.8 & 13.3 & 47.5 & 46.4 \\ \midrule
\multicolumn{11}{c}{\textbf{S1K}} \\
\multirow{1}{*}{H-CoT}  
                        & -                                  & 81.3 & 71.6 & 37.1 & 58.2 & 34.8 & 36.0 & 13.3 & 50.0 & 47.8 \\ \midrule
\multirow{4}{*}{D-CoT}  
                        & \texttt{Qwen2.5-Math-72b-Instruct} & 90.4 & \textbf{75.2} & \textbf{39.3} & 62.9 & 35.3 & \textbf{41.0} & 10.0 & 47.5 & 50.2 \\
                        & \texttt{Qwen-Distil-32B}           & 82.8 & 72.7 & 37.1 & 59.2 & 35.4 & 35.0 & \textbf{16.7} & 50.0 & 48.6 \\
                        & \texttt{Gemini-Flash-2.5}          & 83.1 & 71.4 & 33.8 & 58.2 & 33.2 & 37.8 & 13.3 & 52.5 & 47.9 \\
                        & \texttt{Deepseek-R1}               & 87.0 & 71.4 & 37.9 & 61.0 & 34.5 & 38.5 & \textbf{16.7} & 55.0 & 50.6 \\ \midrule
\multirow{2}{*}{S-CoT}  & \texttt{Self-Refine}               & 87.5 & 71.2 & 34.2 & 62.3 & 35.0 & 40.1 & 16.7 & 60.0 & 50.9 \\
                        & \texttt{Best-of-N}                 & 88.8 & 74.6 & 34.9 & 61.8 & 35.7 & 38.9 & 10.0 & 50.0 & 49.3 \\ \midrule
\rowcolor{gray!20}
\ourmethod
                        & \texttt{w/o Ground Truth}                & 88.9 & 74.0 & 38.2 &  60.5 & \textbf{36.0} & 38.6 & \textbf{16.7} & 57.5 & 51.3 \\ 
\rowcolor{gray!20}
\ourmethod
                        & \texttt{w/ Ground Truth}                & \textbf{90.8} & \textbf{75.2} & \textbf{39.3} &	\textbf{64.4} & \textbf{36.0} & 40.8 & \textbf{16.7} & \textbf{62.5} & \textbf{53.2} \\ \midrule
\multicolumn{11}{c}{\textbf{LIMO}} \\
\multirow{1}{*}{H-CoT}  
                        & -                                 & 78.6 & 67.8 & 34.2 & 57.7 & 33.9 & 34.1 & 10.0 & 50.0 & 45.8 \\ \midrule
\multirow{4}{*}{D-CoT} 
                        
                        & \texttt{Qwen2.5-Math-72b-Instruct} & 82.7 & 73.8 & 37.9 & 57.1 & 34.4 & 37.0 & 13.3 & 50.0 & 48.3 \\ 
                        & \texttt{Qwen-Distil-32B}           & 83.1 & 73.8 & 35.7 & 59.7 & 34.8 & 36.4 & 13.3 & 57.5 & 49.3 \\
                        & \texttt{Gemini-Flash-2.5}          & 83.1 & 71.2 & 37.9 & 58.2 & 34.4 & 35.9 &  6.7 & 57.5 & 48.1 \\
                        & \texttt{Deepseek-R1}               & 83.2 & 73.8 & 36.4 & 58.7 & 35.2 & 34.4 & \textbf{16.7} & 55.0 & 49.2  \\ \midrule
\multirow{2}{*}{S-CoT}  
                        & \texttt{Self-Refine}               & 90.4 & 73.8 & 35.7 & \textbf{64.9} & 34.5 & 40.7 & 10.0 & 55.0 & 50.6 \\
                        & \texttt{Best-of-N}                 & 88.3 & 73.0 & 36.4 & 59.7 & 34.8 & 38.5 & \textbf{16.7} & 52.5 & 50.0                         \\ \midrule

\rowcolor{gray!20}
\ourmethod
                        & \texttt{w/o Ground Truth}                & 87.4 & 73.4 & \textbf{39.0} & 62.3 & 35.4 & 39.7 & 13.3 & \textbf{60.0} & 51.3   \\
\rowcolor{gray!20}                        
\ourmethod
                        & \texttt{w/ Ground Truth}                & \textbf{90.8} & \textbf{76.6} & 38.2 & 62.3 & \textbf{36.1} & \textbf{41.2} & \textbf{16.7} & \textbf{60.0} & \textbf{52.7}   \\
\bottomrule
\end{tabular}
}
\vspace{-0.2cm}
\end{table*}

\paragraph{Dataset.} For training datasets, we select four representative datasets including \texttt{S1K}~\cite{s1k}, \texttt{LIMO}~\cite{limo}, \texttt{Math}~\cite{math}, and \texttt{DeepMath103k}~\cite{deepmath103k}, to span mathematical problems across different scales and complexities. For evaluation datasets, we benchmark \ourmethod on diverse mathematical reasoning datasets, including: (1) Competition and olympiad-level benchmarks, including \texttt{AMC23}~\cite{amc23}, \texttt{MATH500}~\cite{math500}, \texttt{AIME24}~\cite{aime24},
\texttt{Minerva Math}~\cite{minerva_math}
and \texttt{OlympiadBench}~\cite{olympiadbench};
(2) College-level benchmarks, including \texttt{CollegeMath}~\cite{College_math}; and (3) Out-of-domain evaluations, including \texttt{GaoKao-EN 2023}~\cite{gaokaoen23}.
We additionally report results on \texttt{GSM8K}~\cite{best_of_n} for comparison with prior work. Unless otherwise stated, all evaluations use zero-shot prompts with greedy decoding, and we report the Pass@1 metric as defined in the SimpleRL framework~\cite{simplerl}.





\paragraph{Baselines.} 
We benchmark \ourmethod against two representative categories of baseline:  
(1) \textbf{Distilled Chain-of-Thought (D-CoT)} methods, which distill high-quality CoT data from stronger LLMs, including proprietary models (e.g., DeepSeek-R1 ~\cite{deepseek-r1}, Gemini-2.5-Flash \cite{gemini}) and open-source models (e.g., Qwen2.5-Math-72B-Instruct \cite{qwen25mathtechnicalreportmathematical}, DeepSeek-R1-Distill-Qwen-32B \cite{deepseek-r1}). (2) \textbf{Self-Synthesized Chain-of-Thought (S-CoT)} methods, which generate CoT using the model’s own reasoning capability without external models. We consider two representative methods: (i) Best-of-N, which generates multiple reasoning paths and selects the best using a reward model; and (ii) Self-Refine, which iteratively improves initial CoT through feedback-based refinement.  
We also include the original \textbf{human-annotated CoT (H-CoT)} data from given training datasets, which serve as an additional comparison.

\paragraph{Implementation Details.} We adopt the Qwen2.5-7B-Instruct~\cite{qwen25mathtechnicalreportmathematical} as our base model, a widely used general-purpose LLM. To balance generation quality and efficiency, we initialize a population of $N_{\text{pop}}=4$ candidates per problem and perform $t=3$ evolutionary iterations, selecting $k=2$ parents via a fitness function. We set the sampling temperature to 0.6 to encourage output diversity, cap the output length at 2048 tokens, and utilize the vLLM framework~\cite{vllm} to accelerate inference. For cases where the model fails to generate correct answers, we progressively lower the temperature to narrow the search. If no correct solution emerges after evolution, we replace it with D-CoT\footnote{This fallback is rarely used (6.8\% on average).}. For mutation stage, we set mutation temperature $\tau_0=0.6$ and $\lambda=5$. During training, we perform supervised fine-tuning using the OpenRLHF framework~\cite{hu2024openrlhf}. For all approaches, we conduct the hyperparameter search and adop configurations that yield the best performance. See the Appendix for more details.

\begin{figure*}[t]
  \centering
  \includegraphics[width=1.0\textwidth]{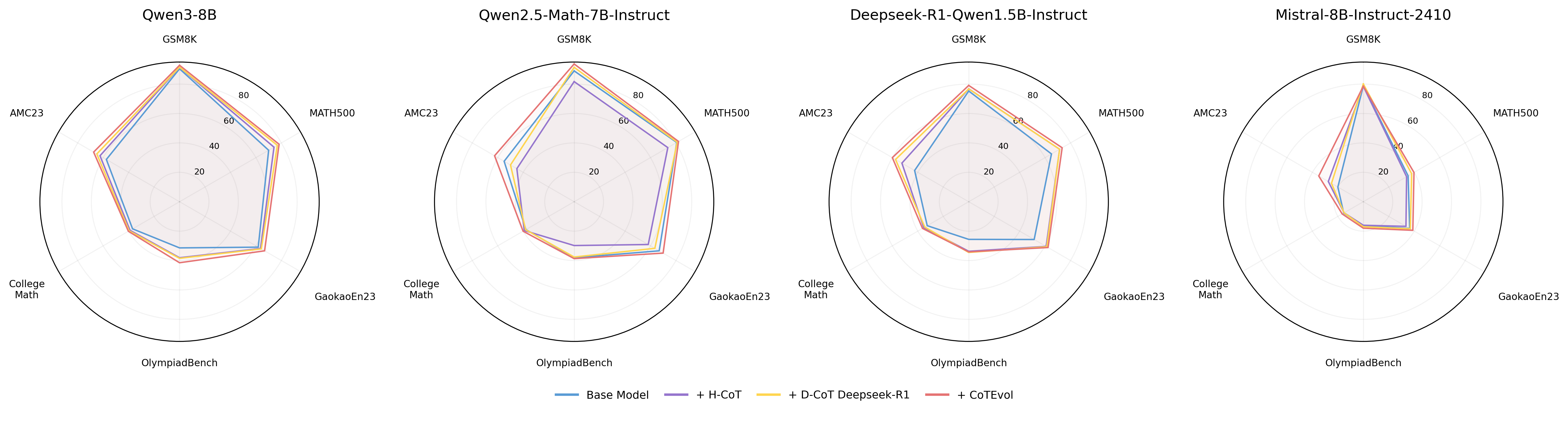}  
  \vspace{-0.8cm}
  \caption{Peformance comparison on math datasets across different models including Qwen3-8B, Qwen2.5-Math-7B-Instruct, Deepseek-R1-Qwen1.5B-Instruct and Mistral-8B-Instruct-2410. 
  }
  \vspace{-0.2cm}
  \label{fig:diff_model}
\end{figure*}

\subsection{Main Results}

Table~\ref{tab:main_result} summarizes the performance of various methods across multiple mathematical benchmarks
on S1K and LIMO. Overall, \ourmethod consistently outperforms all baselines. Compared with the strongest D-CoT (DeepSeek-R1) and S-CoT (Self-Refine) approaches, CoTEvol achieves gains of +2.6 and +2.3 points on S1K, and +3.5 and +2.1 points on LIMO, respectively. These results indicate that the synthesized CoT data effectively activates the model’s latent reasoning capabilities. From a task-level perspective,  \ourmethod yields consistent improvements across difficulty tiers, with larger gains on foundational and intermediate benchmarks and stable gains on competition-level tasks. This trend suggests that while the magnitude of improvement may depend on the base model’s capabilities, \ourmethod remains effective across varying levels of problem complexity. Moreover, although the models are fine-tuned for math data, evaluations on out-of-distribution benchmarks show minimal performance degradation, suggesting that the observed improvements do not compromise generalization, as reported in Appendix.

\begin{figure}[]
  \centering
  \includegraphics[width=0.38\textwidth]{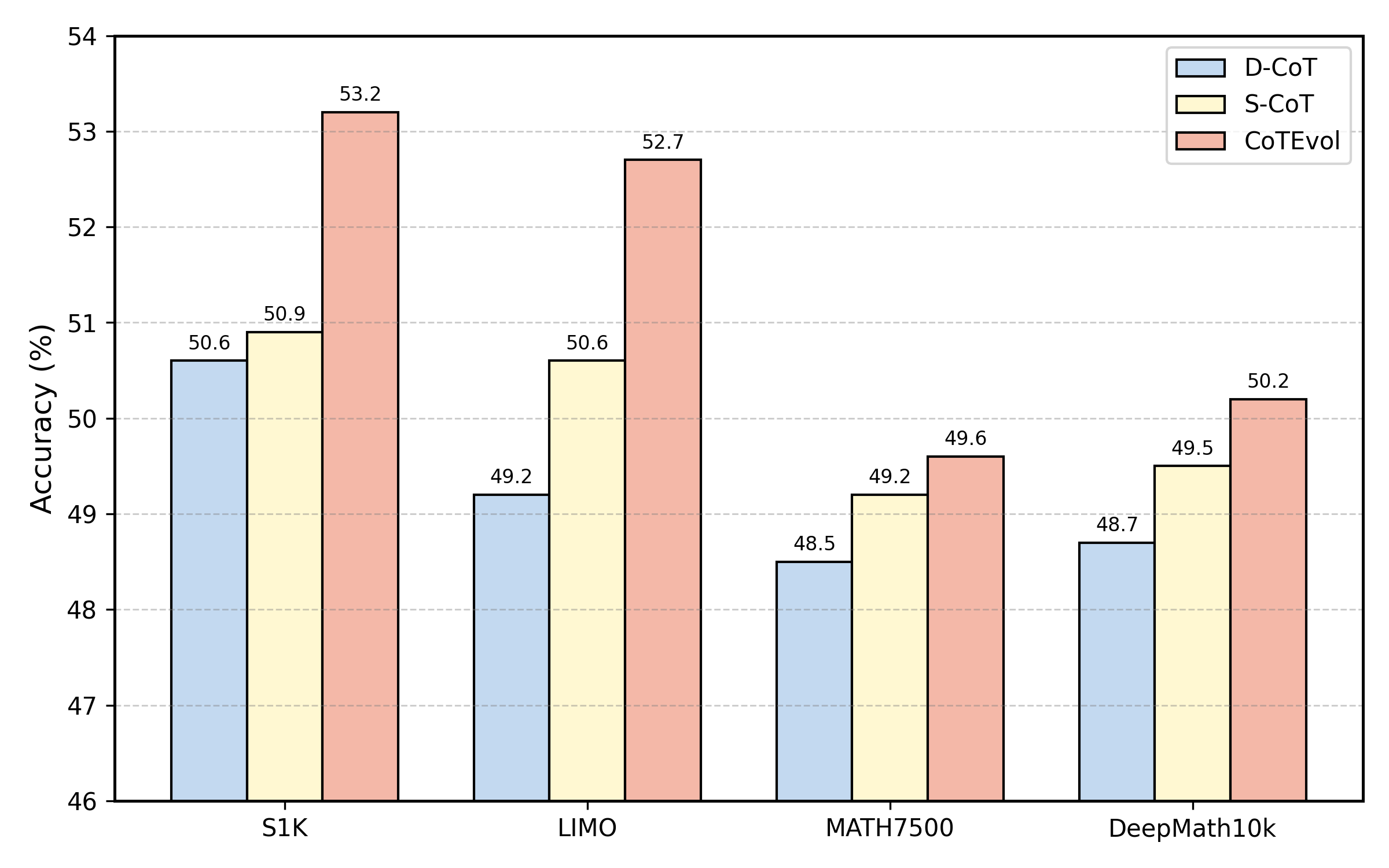}  
  \vspace{-0.15cm}
  \caption{Performance comparison of D-CoT, S-CoT and \ourmethod across different data scales.
  }
  \vspace{-0.5cm}
  \label{fig:data_sacle}
  
\end{figure}

\paragraph{\ourmethod with different models.} 
We evaluate the generality of \ourmethod across four models with diverse scales and architectures: a recent model (Qwen3-8B~\cite{qwen25mathtechnicalreportmathematical}), a mathematics-specialized model (Qwen2.5-Math-7B-Instruct~\cite{qwen25mathtechnicalreportmathematical}), a reasoning-distilled model (Deepseek-R1-Qwen1.5B-Instruct~\cite{deepseek-r1}), and a general-purpose model (Mistral-8B-Instruct-2410~\cite{ministral8b2024}). 
As shown in Figure~\ref{fig:diff_model}, \ourmethod consistently outperforms all baselines across models, demonstrating strong model-agnostic effectiveness. Notably, larger performance gains are observed for mathematically stronger models, suggesting that \ourmethod generates high-quality CoTs that better activate pretrained reasoning capacity.

\paragraph{\ourmethod with different scale datasets.} To evaluate the scalability of \ourmethod, we compare with S-CoT and D-CoT distilled from DeepSeek-R1 on four datasets ranging from 1K to 10K samples, as detailed in Figure~\ref{fig:data_sacle}. The results demonstrate that \ourmethod can evolve effectively across datasets of varying scales, consistently achieving strong performance. Notably, we observe that merely increasing the volume of training data does not necessarily improve accuracy. Instead, data diversity and quality play a more decisive role, as also observed in~\citet{limo}. Consequently, we focus our subsequent experiments on S1K and LIMO, two carefully curated and high-quality datasets.

\subsection{Further Analysis}

\paragraph{Ablation Study.}
We conduct ablation experiments on the S1K dataset to evaluate the contribution of each component in \ourmethod. Table~\ref{tab:ablation_small} reports the average performance across benchmarks. Removing any component leads to a consistent performance drop, demonstrating their complementary contributions. Specifically, eliminating either the fitness function or crossover leads to a similar performance drop, highlighting the importance of both effective selection and reflective recombination. In contrast, removing mutation results in a smaller yet consistent decrease, indicating its role in fine-grained local exploration. Overall, these components jointly enable effective evolutionary synthesis of high-quality CoT data. Detailed results and case study are provided in the Appendix.

\begin{table}[!t]
\centering
\small
\setlength{\tabcolsep}{3.7mm}
\renewcommand\arraystretch{1.15}
\vspace{-0.2cm}
\caption{Ablation experiments trained on S1K dataset. Detailed results arcoss all benchmarks is in Appendix.}
\label{tab:ablation_summary} \label{tab:ablation_small}
\begin{tabular}{cccc}
\toprule
\multicolumn{3}{c}{\textbf{Component}} & \multirow{2}{*}{\textbf{AVG.}} \\
\cmidrule(lr){1-3}
\textbf{Crossover} & \textbf{Mutation} & \textbf{Fitness} &  \\
\midrule
           & \checkmark & \checkmark & 51.2 \\
\checkmark &            & \checkmark & 52.0 \\
\checkmark & \checkmark &            & 51.3 \\
\checkmark & \checkmark & \checkmark & \textbf{53.2} \\
\bottomrule
\end{tabular}
\vspace{-0.3cm}
\end{table}

\paragraph{Effect of Ground Truth.} 
We evaluate the effectiveness of CoTEvol without ground-truth (GT) supervision from two perspectives: evolutionary success rate and downstream generalization performance. As shown in Table~\ref{tab:main_result} and Table~\ref{tab:dataset_sr_comparison}, removing GT supervision leads to a slight drop in evolutionary success rate. However, when the resulting reasoning trajectories are directly used for supervised fine-tuning without any additional distillation data, they still yield consistent improvements over the base model on both S1K and LIMO.

We attribute this robustness to two factors: (i) the model’s ability to act as a qualitative verifier during crossover, where evaluating logical consistency is easier than generation, and (ii) the use of step-level entropy in mutation as an unsupervised signal to identify and refine uncertain reasoning steps. Together, these mechanisms enable effective optimization even without explicit correctness feedback signal.

\begin{table}[t!]
\setlength{\tabcolsep}{0.7mm}  
\renewcommand\arraystretch{1.3} 
\caption{Evolutionary success rates (SR) across S1K and LIMO datasets.}\label{tab:dataset_sr_comparison}
\vspace{-5pt}
\resizebox{\linewidth}{!}{
    \begin{tabular}{l|ccc} 
    \toprule
    \textbf{Dataset} & \textbf{Original SR} & \textbf{Evol. SR (w/o GT)} & \textbf{Evol. SR (w/ GT)} \\ \midrule
    S1K  & 0.359 & 0.491 & 0.825 \\
    LIMO & 0.420 & 0.537 & 0.881 \\
    \bottomrule
    \end{tabular}
}
\end{table}

\begin{table}[t!]
\setlength{\tabcolsep}{0.7mm}  
\renewcommand\arraystretch{1.3} 
\caption{Evolutionary success rates (SR) across different difficulty tiers.}\label{tab:difficulty_sr}
\vspace{-5pt}
\resizebox{\linewidth}{!}{
    \begin{tabular}{l|ccc}
    \toprule
    \textbf{Difficulty Tier} & \textbf{Original SR} & \textbf{Evol. SR (w/o GT)} & \textbf{Evol. SR (w/ GT)} \\ \midrule
    Simple-500   & 0.899 & 0.910 & 0.910 \\
    Complex-500  & 0.755 & 0.820 & 0.852 \\ 
    Advanced-500 & 0.405 & 0.450 & 0.622 \\
    \bottomrule
    \end{tabular}
}
\vspace{-0.3cm}
\end{table}

\paragraph{Impact of Problem Difficulty on Evolutionary Performance.} We observe that problem difficulty plays a critical role in determining the effectiveness of Co
TEvol. To analyze this, we partition the dataset into three difficulty tiers following the LIMO setting\cite{limo}: Simple-500, Complex-500 and Advanced-500. We evaluate the success rate after three rounds of evolution under both search (without ground-truth supervision) and GT-guided (with ground-truth supervision) settings. As shown in Table~\ref{tab:difficulty_sr}, in the Simple-500 setting, performance quickly saturates near 90\%, leaving minimal room for logical refinement. In contrast, in the Advanced-500 setting, performance is limited by the model’s capability, resulting in low-quality initial reasoning trajectories and hindering the evolutionary process at early stages. Notably, the Complex-500 setting provides a more favorable condition, where initial solutions are informative yet imperfect, enabling the evolutionary process to iteratively refine reasoning trajectories. These results suggest that CoTEvol is most effective under moderate difficulty, where both sufficient optimization space and viable initial candidates are present.

\begin{figure}[t]
  \centering 
  \begin{subfigure}[b]{0.235\textwidth}
    \centering
    \includegraphics[width=\textwidth]{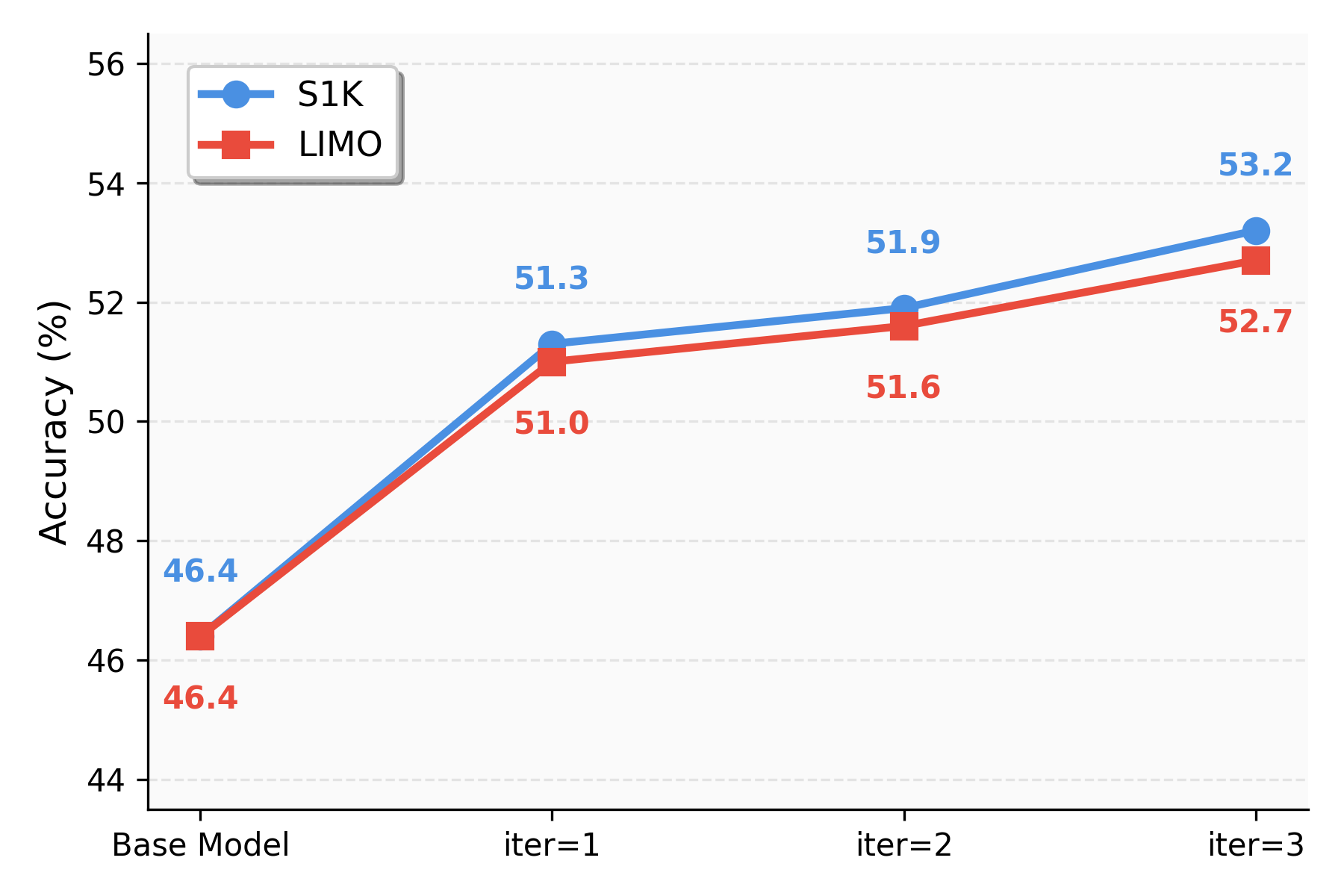}
    \vspace{-0.6cm}
    \caption{}
    \label{fig:iter}
  \end{subfigure}
  \hfill
  \begin{subfigure}[b]{0.235\textwidth}
    \centering
    \includegraphics[width=\textwidth]{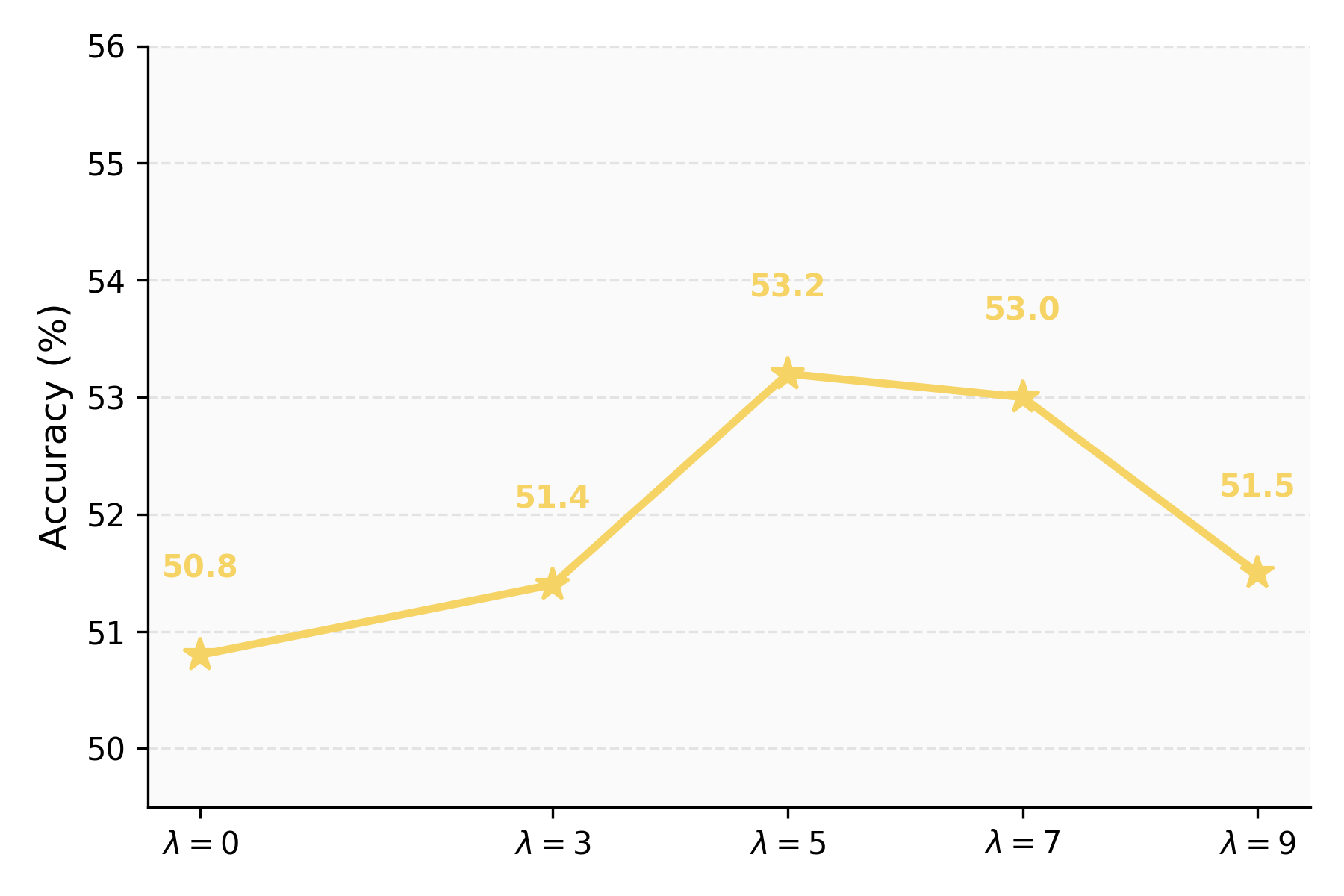}
    \vspace{-0.6cm}
    \caption{}
    \label{fig:lambda}
  \end{subfigure}
  \vspace{-0.2cm}
  \caption{(a) Comparison of iteration rounds; (b) impact of $\lambda$ on model performance.}
  \vspace{-0.2cm}
  \label{fig:iter_lambda}
\end{figure}

\paragraph{Effect of different iteration rounds.} We analyze the effect of evolutionary rounds on model performance using the S1K and LIMO datasets. As shown in Figure~\ref{fig:iter}, performance consistently improves as the number of evolutionary rounds increases, reaching gains of 6.8 and 6.3 points on S1K and LIMO at $iter=3$, respectively. The largest improvement occurs in the first round, indicating that early evolution rapidly enhances reasoning quality, while additional rounds mainly refine and stabilize performance with diminishing returns. 

\paragraph{Effect of hyperparameter $\lambda$.} Figure~\ref{fig:lambda} presents the effect of $\lambda$ in uncertainty-guided mutation. We find that as $\lambda$ increases, higher mutation temperature enables broader exploration and helps escape uncertain reasoning steps. Performance peaks at $\lambda=5$, which balances exploration and stability. Further increasing $\lambda$ leads to overly aggressive exploration, enlarging the search space and reducing performance. We thus set $\lambda=5$ in all experiments.


\begin{figure}[!t]
  \centering
  \includegraphics[width=0.38\textwidth]{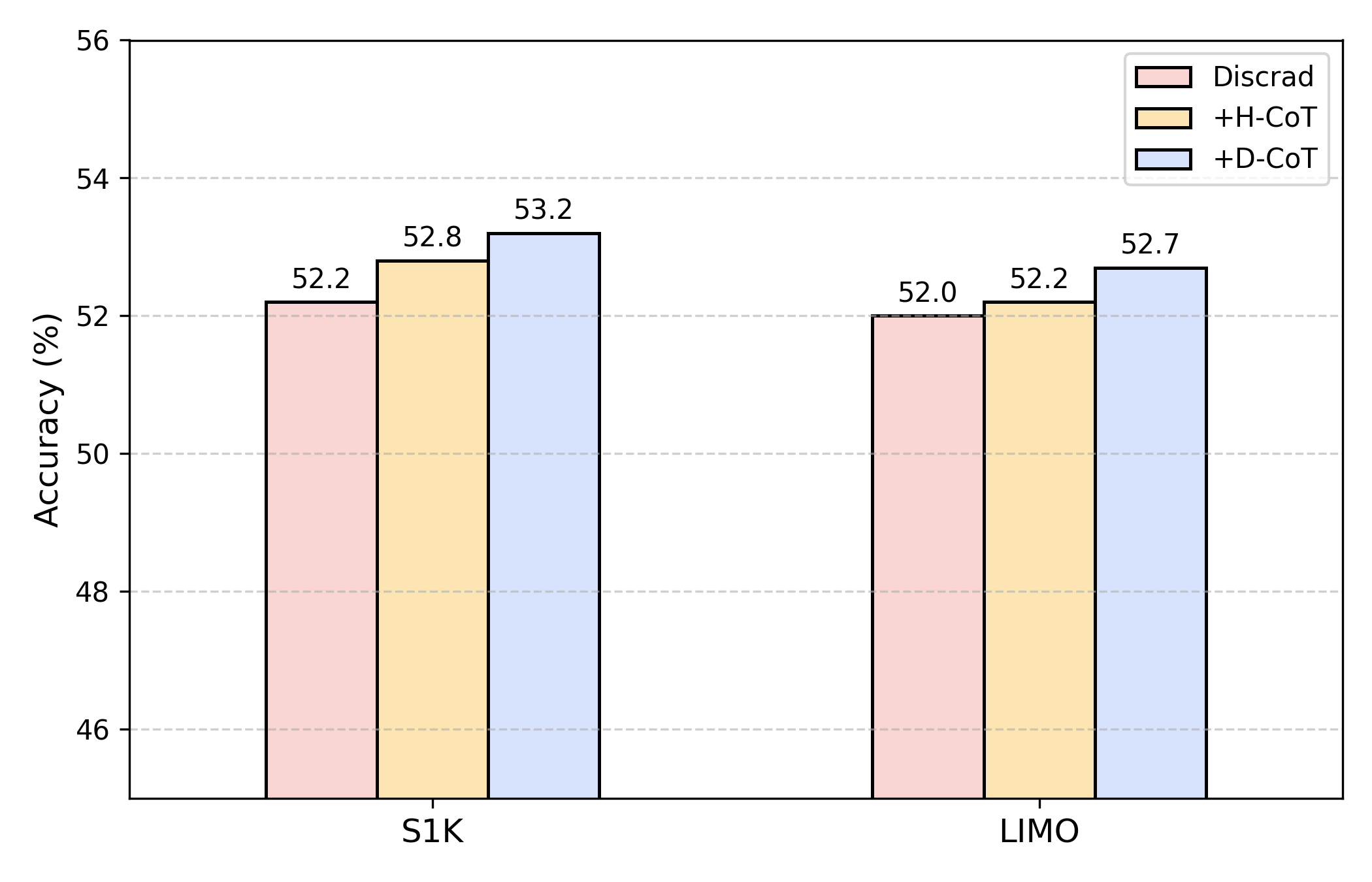}  
  \vspace{-0.3cm}
  \caption{Comparison of different data fusion strategies using Qwen2.5-7B-Instruct on S1K and LIMO datasets.
  }
  \vspace{-0.5cm}
  \label{fig:mix}
\end{figure}

\paragraph{Mixed Data Strategy.} Figure~\ref{fig:small_pic} shows that \ourmethod consistently achieves over 80\% problem-solving accuracy, highlighting the effectiveness of \ourmethod. For samples that fail to evolve, we explore three strategies: (1) discarding failed samples, (2) substituting them with manually annotated H-CoT, and (3) replacing them with distilled D-CoT. As illustrated in Figure~\ref{fig:mix}, the hybrid D-CoT strategy delivers the best performance, benefiting from the strong reasoning ability of frontier LLMs on complex problems. On LIMO, however, gains are limited since the training-stage success rate is already high and fewer failures need to be patched. These results suggest a practical synthesis paradigm: generate most data via \ourmethod, and selectively patch failures with high-quality distilled traces. This hybrid approach striking the balance between quality and cost, making it well-suited for real-world deployment.
\paragraph{Effect of Test-Time Methods during Inference.} we compare our method with Best-of-N and Self-Refine in terms of their effectiveness in searching the solution space during inference, as shown in Table~\ref{tab3-test-time methods}. Best-of-N achieves relatively strong performance by leveraging repeated sampling and an explicit reward signal. In contrast, Self-Refine yields only limited improvements, as it operates on a single reasoning trajectory through iterative critique and refinement. Our method, by contrast, combines global recombination with local mutation, enabling more thorough exploration of the solution space and a higher likelihood of reaching correct solutions. These results demonstrate that our approach remains effective even during inference.


\begin{table}[!t]
\centering
\small
\renewcommand\arraystretch{1.3}
\caption{Comparison of different test-time search methods during inference under the Pass@10 metric.}
\label{tab3-test-time methods}
\begin{tabular*}{\linewidth}{@{\extracolsep{\fill}}lcc}
\toprule
\textbf{Method} & \textbf{MATH500} & \textbf{AMC23} \\
\midrule
Qwen2.5-7B-Instruct & 69.4 & 47.4 \\
\ + Best-of-N & 85.4 & 82.5 \\
\ + Self-Refine & 85.4 & 85.0 \\
\ + \ourmethod & \textbf{86.2} & \textbf{87.5} \\
\bottomrule
\end{tabular*}
\end{table}

\begin{table}[!t]
\centering
\caption{Comparison of different synthesis strategies in terms of FLOPs (values $\times 10^{12}$).}
\vspace{-0.1cm}
\label{tab:flops}
\resizebox{\linewidth}{!}{
\begin{tabular}{lccc}
\toprule
Method & Best-of-N & Self-Refine & \ourmethod \\
\midrule
FLOPs  & 1689.53 & 733.95 & 453.83 \\

\bottomrule
\end{tabular}
}
\vspace{-0.2cm}
\end{table}

\paragraph{Efficiency Comparison of Synthesis Strategies.} We compare the computational cost of different reasoning synthesis strategies in terms of FLOPs (Table~\ref{tab:flops}). Both Best-of-N and Self-Refine generate 10 reasoning paths using \texttt{Qwen2.5-7B-Instruct}, whereas \ourmethod begins with 4 initial solutions and produces 10 candidates through three iterations. Best-of-N incurs the highest cost due to the additional \texttt{Qwen2.5-Math-RM-72B} reward model. Self-Refine reduces generation cost but still performs critique and refinement on all solutions. In contrast, \ourmethod maintains low computational cost by applying crossover and mutation only to the top 2 solutions selected via fitness functions. Despite three iterations, the combination of a small initial population and precise selection significantly reduces FLOPs. Furthermore, as RL provides an alternative approach for CoT generation, we also include a comparison in the Appendix.

Additional ablation studies, including out-of-distribution performance, CoT quality analysis, the impact of trajectory quantity, case studies, and the effect of different initialization seeds on robustness, are provided in the Appendix.



\section{Conclusion}
This paper presents \ourmethod, a genetic algorithm-based reasoning framework for the automatic synthesis of high-quality chain-of-thought data in mathematical reasoning. Our method formulates reasoning as a structured search problem and integrates fitness-guided selection with reflective global recombination and uncertainty-aware local mutation, enabling effective exploration of the solution space. Experimental results demonstrate that \ourmethod consistently outperforms prior approaches across multiple benchmarks and model scales. These results underscore the potential of evolutionary search as a scalable and principled framework, paving the way for further improvements in reasoning-centric language models.

\section{Limitations}
Although our evolutionary framework is able to generate structurally diverse and effective reasoning trajectories, our current supervised fine-tuning (SFT) strategy primarily relies on selecting a single highest-fitness solution per problem. While this best-only strategy provides strong and stable supervision signals, it does not fully exploit the diversity of evolved trajectories. We further explore multi-trajectory training, but empirically find that naively incorporating multiple solutions does not outperform the best-only setting, suggesting that effectively leveraging diverse reasoning paths remains a challenging problem. This indicates that the value of evolutionary diversity lies not only in generation, but also in developing more effective mechanisms for aggregation and utilization.

\clearpage
\appendix

\section{Appendix}
\label{sec:appendix}
\subsection*{Dataset Details} 
We provide an overview of the datasets used in our training and evaluation pipelines. The training datasets include S1K, LIMO, MATH, and DeepMath103k, covering a wide range of math problem types and quantities. Notably, both S1K and LIMO are carefully curated benchmark datasets. From the larger MATH dataset, we select a representative subset of 7.5k examples, while for DeepMath103k, we sample 10k problems stratified by difficulty. We leverage this setup to assess the impact of dataset scale and complexity on the effectiveness of our method. For evaluation, we select diverse benchmarks including GSM8K, MATH500, Minerva Math, GaoKao-EN 2023, OlympiadBench, AMC23 and AIME24. This diversity ensures a comprehensive assessment of model capability across difficulty and domain boundaries. Table~\ref{tab:data_statistics} summarizes the number of samples, difficulty levels,  data sources and description for each dataset. 
\vspace{-0.2cm}


\subsection*{Implementation Details}
For our method, we mainly adopt Qwen2.5-7B-Instruction as the base model, which is a widely used and general-purpose LLM. For each question in the training set, we initialize a population of 4 candidate solutions and perform 3 rounds of self-evolution. Each iteration produces two new offspring: one via crossover and the other via mutation. We set the batch size to 32, use a sampling temperature of 0.6 for both response generation and variation operations, and cap the maximum output length at 2048 tokens. Furthermore, we set $C_{min}=0.5$, $C_{max}=1.0$, $W_{min}=1.0$ and $W_{max}=0.5$ for length-based reward verifier. During mutation, we set $\lambda=5$ to improve temperature to enhance exploration. Finally, we obtain a population of 8 candidate solutions and select the one with the highest fitness score as the final solution. 

For data distillation methods (D-CoT), we adopt a two-stage distillation strategy. Specifically, we first set the temperature to 0.6 and generate five responses for each question using the target model (eg. Qwen2.5-Math-72B-Instruct). If any of the responses produce the correct final answer (verified against the ground truth), we retain the corresponding reasoning trace as a valid distilled CoT. If none of the responses are correct, we fall back to the Deepseek-R1 solution and prompt the model to rewrite it in its style to generate a valid CoT data. 

For self-generated methods via test-time computation (i.e., Best-of-N and Self-Refine), we adopt a standardized computation budget based on the maximum number of output tokens to ensure fair comparison. Specifically, we constrain the total number of generated tokens to 16,384 (i.e., 8 × 2048). In the Best-of-N setting, we independently generate eight candidate responses and select the highest quality solution among the correct ones using the Qwen2.5-Math-72B-RM reward model. For Self-Refine, we generate four initial responses and apply two iterations of refinement, prioritizing correction of erroneous solutions, followed by selection of the best correct response using the same reward model. Figure 1 illustrates the success rates of different methods on the S1K and LIMO benchmarks. For incorrect problems, we also utilize distilled data from Deepseek-R1.

All experiments for \ourmethod and baselines are conducted using PyTorch 2.5.1 and CUDA 12.4 with vLLM version 0.6.4 for efficient batched decoding on 4×NVIDIA H100 GPUs with 80G memory. During the SFT training stage of each method, we experiment with batch sizes of 16 and learning rates of 1e-5, 5e-6, 7e-7, with the number of training epochs set to 1 and 3. During the testing stage, we set the temperature to 0.0 and conduct a hyperparameter search to select the configuration that yields the best performance. The code is provided in the supplementary materials and will be made publicly available in the near future.

\begin{table*}[t]
\centering
\fontsize{7.5}{10.5}\selectfont 
\setlength{\tabcolsep}{1.2mm}
\renewcommand\arraystretch{1.3}
\caption{Statistics, sources, and descriptions of training and evaluation datasets.}
\label{tab:data_statistics}
\begin{tabular}{cccccl}
\toprule
\textbf{Datasets} & \textbf{Split} & \textbf{Samples} & \textbf{Difficulty} & \textbf{Source} & \textbf{Description} \\
\midrule
S1K & Train & 1,000 & Hard & \cite{s1k} & A curated math dataset with structured reasoning paths. \\
LIMO & Train & 817 & Medium & \cite{limo} & A high-quality reasoning dataset with rigorous reasoning chains. \\
MATH & Train & 7,500 & Hard & \cite{math} & Benchmark with 12.5k high-school competition problems. \\
DeepMath103k & Train & 10,000 & Hard & \cite{deepmath103k} & A competition-level dataset spanning difficulties from Level 1 to Level 10. \\
GSM8K & Test & 1,319 & Easy & \cite{best_of_n} & Grade school-level problems requiring basic arithmetic and logic. \\
MATH500  & Test & 500 & Medium & \cite{math500} & Hard competition-style problems from the MATH benchmark. \\
Minerva Math & Test & 272 & Hard & \cite{minerva_math} & High-school to undergrad-level problems sources from Google's corpus. \\
GaoKao-EN 2023 & Test & 385 & Medium & \cite{gaokaoen23} & English-translated Chinese National College Entrance Examination\\
OlympiadBench & Test & 675 & Hard & \cite{olympiadbench} & High-level math olympiad problems requiring deep reasoning. \\
CollegeMath & Test & 2,818 & Medium & \cite{College_math} & Math problems from college-level exams. \\
AMC23 & Test & 40 & Hard & \cite{amc23} & Problems from the 2023 AMC math contest. \\
AIME24 & Test & 30 & Hard & \cite{aime24} & Problems from the 2024 AIME I and II math contest. \\
\bottomrule
\end{tabular}
\end{table*}

\subsection*{Compared with Mind Evolution} 
In this section, we highlight three key differences between our approach and Mind Evolution: 
\textbf{(1) task differences.} Mind Evolution primarily focuses on planning tasks, such as TravelPlanner, Trip Planning, and Meeting Planning. In contrast, our method targets mathematical reasoning, which presents fundamentally different challenges and objectives. 
\textbf{(2) goal differences.} Our approach is designed to synthesize high-quality CoT data, aiming to enhance both the quality and diversity of generated solutions. This data is then used to improve model accuracy and generalization through supervised fine-tuning (SFT). Mind Evolution, on the other hand, is a test-time computation that operates only during inference and does not involve any model parameter updates. 
\textbf{(3) differences in module design.} For the fitness function, we incorporate multiple verifiers, including an answer correctness verifier, a format matching verifier, and a length-based reward verifier, which jointly assess solutions from the perspectives of correctness, formatting consistency, and response length. 
In contrast, Mind Evolution relies on strict constraint checking, applying penalties and feedback when conditions are violated. For the genetic operations design, we propose reflective global crossover and uncertainty-guided local mutation to effectively use self-reflective ability and and uncertainty estimates to perform fine-grained optimization over reasoning paths. This reflects the classical evolutionary mechanism, with crossover enabling global structural recombination and mutation introducing targeted local variations. \textbf{(4) different challenges.} Mind Evolution leverages the significantly larger Gemini 1.5 Flash model, whereas our approach is based on smaller LLMs of varying scales. While large frontier models can more easily support evolutionary processes due to their capacity and robustness, enabling self-evolution in smaller models presents a substantially greater challenge. However, in real-world scenarios where computational resources are limited and application-specific constraints exist, our method offers a more practical and deployable solution.
Overall, while both approaches adopt an evolutionary framework, our method is different from Mind Evolution in both problem setting and methodological goals.

\subsection*{Prompt Details}\label{sec:prompt_details}
During the evolutionary process, we design a set of specialized prompts to explicitly guide different stages of solution generation and refinement.
Specifically, the framework includes a base response prompt for initial solution generation, a set of crossover prompts for feedback-driven recombination, and a set of mutation prompts for uncertainty-guided exploration. 
\textbf{For the crossover stage}, we first generate structured reflection-based feedback according to the correctness patterns of the selected parent CoTs.
Three distinct feedback-generation prompts are employed, corresponding to \emph{Elite Merging}, \emph{Success--Error Fusion}, and \emph{Failure Pattern Summary}, which handle the cases where both parents are correct, only one parent is correct, or both parents are incorrect, respectively.
The resulting feedback is then injected into a dedicated critique-and-refine prompt, which guides the LLM to perform crossover and synthesize a new offspring CoT that integrates high-quality reasoning patterns while avoiding known errors. \textbf{For the mutation stage}, we employ two refinement prompts. One performs localized mutation from the reasoning step with highest uncertainty, identified via step-level entropy, encouraging alternative trajectories while preserving preceding stable steps. The other performs global mutation by regenerating the entire CoT from the beginning, targeting the step with highest uncertainty, which corresponds to the initial reasoning step. Finally, to evaluate the quality of the evolved reasoning traces, we prompt GPT-4.1 to perform pairwise comparison between the evolved CoT from \ourmethod and the distilled CoT, selecting the preferred solution based on overall reasoning quality. All prompts is shown in Appendix Figure~8-14.


\begin{table*}[ht]
\setlength{\tabcolsep}{0.7mm}  
\renewcommand\arraystretch{1.3}
\caption{Ablation experiments across different mathematical benchmarks trained on S1K dataset.}\label{tab:ablation}
\vspace{-5pt}
\resizebox{\linewidth}{!}{
    \begin{tabular}{ccc|ccccccccc}
    \toprule
    \multicolumn{3}{c|}{\textbf{Component}} & \multicolumn{8}{c}{\textbf{Benchmarks}} \\ 
    \texttt{Cros.} & \texttt{Mut.} & \texttt{Fit.} &  \textbf{GSM8k} & \textbf{MATH500} & \textbf{MinervaMath} & \textbf{GaokaoEn23} & \textbf{OlympiadBench} & \textbf{CollegeMath} & \textbf{AIME24} & \textbf{AMC23} & \cellcolor{gray!20}\textbf{AVG.} \\ \midrule
    & \checkmark & \checkmark & 87.5  & 72.2  & 39.3 & 62.6 & 35.3 & 39.4 & 16.7 & 53.5 & \cellcolor{gray!20}51.2 \\
    \checkmark & & \checkmark & 88.6  & 74.2  & 38.6 & 63.9 & 35.8 & 39.4 & 13.3 & 62.5 & \cellcolor{gray!20}52.0 \\
    \checkmark & \checkmark & & 88.9 & 74.0 & 38.2 &  60.5 & 36.0 & 38.6 & 16.7 & 57.5 & \cellcolor{gray!20}51.3  \\
    \checkmark & \checkmark & \checkmark &  90.8 & 75.2 & 39.3 &	64.4 & 36.0 & 40.8 & 16.7 & 62.5 & \cellcolor{gray!20}53.2 \\ 
    \bottomrule
    \end{tabular}
}
\vspace{-0.2cm}
\end{table*}
\paragraph{More Ablation Experiments}\mbox{}\\
The ablation results of each component across all datasets are presented in Table~6, where Cross. denotes the reflective global crossover, Mut. denotes the uncertainty-guided local mutation, and Fit. denotes the fitness function.

\begin{table}[!t]
\centering
\caption{Comparison of out-of-distribution generalization across different dataset.}
\label{tab:ood}
\resizebox{\linewidth}{!}{
\begin{tabular}{lccc}
\toprule
Method & ARC-c & GPQA-diamond & MMLU-Pro \\
\midrule
Base  & 90.02 & 31.82 & 56.69 \\
Ours  & 90.58 & 31.82 & 56.41 \\
\bottomrule
\end{tabular}
}
\end{table}

\begin{table}[!t]
\centering
\footnotesize 
\setlength{\tabcolsep}{0pt} 
\renewcommand\arraystretch{1.3}
\caption{Comparison of \ourmethod and D-CoT based on GPT-4.1 and human judgment.}
\label{tab:quality}
\begin{tabular*}{\linewidth}{@{\extracolsep{\fill}}l|ccc|ccc}
\toprule
& \multicolumn{3}{c|}{S1K} & \multicolumn{3}{c}{LIMO}  \\ 
& Win ($\uparrow$) & Tie ($\uparrow$) & Lose ($\downarrow$) & Win ($\uparrow$) & Tie ($\uparrow$) & Lose ($\downarrow$)  \\ 
\midrule
GPT-4.1  & 6 & 50 & 44 & 13 & 47 & 40    \\
Human    & 11 & 52 & 39 & 15 & 52 & 33    \\
\bottomrule
\end{tabular*}
\vspace{-0.2cm}
\end{table}

\textbf{Out-of-Distribution Performance.} Since our training primarily targets mathematical reasoning, we further evaluate the generalization capability of our method on three out-of-distribution (OOD) benchmarks: ARC-c (open-domain reasoning), GPQA-diamond (graduate-level science knowledge), and MMLU-Pro (reasoning-focused questions from academic exams and textbooks). These benchmarks differ substantially from the training distribution in both domain and reasoning style. To mitigate potential answer pattern bias or contamination, we randomly shuffle the multiple-choice options for all evaluations. Results in Table~\ref{tab:ood} show that while our method substantially improves mathematical reasoning performance, it preserves generalization ability on out-of-distribution benchmarks without degradation.

\textbf{Quality Analysis.} We compare the reasoning quality of CoT generated by \ourmethod with D-CoT distilled from DeepSeek-R1 on the S1K and LIMO datasets, using win/tie/loss judgments from both GPT-4.1 and human evaluators. As shown in Table~\ref{tab:quality}, although \ourmethod yields slightly lower average scores, it matches or outperforms D-CoT in over half of the test cases, indicating competitive reasoning quality. Notably, \ourmethod achieves this without relying on external models, instead improving reasoning chains through verifiable answers and genetic search. Detailed reasoning examples of \ourmethod are provided in Figure 15.

\textbf{Comparision with RL method.} Reinforcement Learning (RL), especially recent policy optimization methods such as GRPO, has been widely adopted to improve the reasoning capabilities of LLMs by optimizing policies through reward signals. In contrast to RL-based approaches that require iterative gradient updates and careful hyper-parameter tuning, \ourmethod is a training-free evolutionary search framework. It formulates CoT synthesis as a combinatorial optimization problem and directly evolves reasoning trajectories via crossover and mutation, guided by lightweight and verifiable fitness signals. This design avoids common challenges in RL, including training instability and entropy collapse, while enabling stable convergence within three iterations. We further conduct a comparison between \ourmethod and GRPO on the DeepSeek-R1-Distill-Qwen-1.5B model using the S1K dataset. Under the same hardware setup (4×A100 GPUs), GRPO requires 4.7 hours of training over three epochs, whereas \ourmethod reaches a higher performance level in only 1.3 hours, reducing the time cost by 72\%. Despite the significantly lower computational overhead, \ourmethod consistently outperforms GRPO in Pass@1 accuracy, achieving 23.3\% vs.\ 21.3\% on AIME24 and 60.0\% vs.\ 58.1\% on AMC23. These results demonstrate that evolutionary search offers a more cost-effective and practical alternative to policy optimization for large-scale CoT synthesis.

\begin{table*}[ht]
\setlength{\tabcolsep}{0.7mm}  
\renewcommand\arraystretch{1.3} 
\caption{Experimental results across different random seeds trained on LIMO dataset.}\label{tab:seed_results}
\vspace{-5pt}
\resizebox{\linewidth}{!}{
    \begin{tabular}{l|ccccccccc}
    \toprule
    \textbf{Seed} & \textbf{GSM8k} & \textbf{MATH500} & \textbf{MinervaMath} & \textbf{GaokaoEn23} & \textbf{OlympiadBench} & \textbf{CollegeMath} & \textbf{AIME24} & \textbf{AMC23} & AVG. \\ \midrule
    seed=42   & 90.8 & 76.6 & 38.2 & 62.3 & 36.1 & 41.2 & 16.7 & 60.0 & 52.7 \\
    seed=1234 & 89.9 & 74.4 & 40.4 & 63.4 & 36.3 & 40.7 & 23.3 & 55.0 & 52.9 \\
    seed=3407 & 90.2 & 74.0 & 39.7 & 64.4 & 35.7 & 40.8 & 16.7 & 55.0 & 52.1 \\
    \bottomrule
    \end{tabular}
}
\vspace{-0.2cm}
\end{table*}

\textbf{Robustness across Different Random Seeds.} To mitigate the potential bias introduced by random seed initialization, we conducted experiments on the LIMO dataset using three distinct seeds (42, 1234, and 3407). The generalization results across various benchmarks are summarized in Table~\ref{tab:seed_results}. The marginal variance observed across different seeds indicates that our approach is largely insensitive to initial random states, thereby confirming the robustness and stability of our method.

\textbf{Impact of Trajectory Quantity.} We further conducted a comparative analysis between a multi-solution setting and the highest-fitness baseline to evaluate the influence of trajectory quantity on model performance. We aggregate all correct reasoning paths per problem instead of exclusively selecting the highest-fitness solution and apply a Rouge-L filter to mitigate redundancy and maintain diversity, resulting in an augmented SFT dataset with an average of 4.655 solutions per problem. Paradoxically, fine-tuning Qwen-2.5-7B-Instruct on this multi-solution dataset led to a performance degradation, with average accuracy declining from 44.7\% to 43.4\%, as shown in Table~\ref{tab:solution_comparison}. This result suggests that not all correct trajectories provide equivalent value as supervisory signals. Specifically, while lower-fitness solutions yield correct final answers, they frequently involve redundant steps or suboptimal logical structures. The inclusion of such paths introduces logical noise, which dilutes the model's ability to learn a clean and efficient reasoning distribution, thereby hindering overall performance.

\begin{table}[ht]
\setlength{\tabcolsep}{0.7mm}  
\renewcommand\arraystretch{1.3} 
\caption{Performance comparison between multi-solutions and single-solution training settings.}\label{tab:solution_comparison}
\vspace{-5pt}
\resizebox{\linewidth}{!}{
    \begin{tabular}{l|ccccc} 
    \toprule
    \textbf{Setting} & \textbf{MinervaMath} & \textbf{GaoKaoEn23} & \textbf{OlympiadBench} & \textbf{CollegeMath} & \textbf{Avg.} \\ \midrule
    Acc (multi-solutions) & 36.8 & 61.0 & 36.7 & 39.2 & 43.4 \\
    Acc (single-solution) & 38.2 & 62.3 & 36.1 & 41.2 & 44.7 \\
    \bottomrule
    \end{tabular}
}
\vspace{-0.2cm}
\end{table}
\subsection*{Case Study}
To more intuitively illustrate the advantages of \ourmethod, we present a case study in Figure~\ref{fig:case_study}. The solution produced by \ourmethod exhibits a highly structured problem-solving process, frequently incorporating reflective cues such as “verify,” which suggest a stronger degree of internal self-checking and deeper reasoning.
As a comparative baseline, we analyze distilled solutions derived from the frontier DeepSeek-R1 model. We observe that even state-of-the-art LLMs may occasionally exhibit logical inconsistencies, where the final answer is correct but intermediate reasoning steps are flawed or insufficiently grounded.
These results indicate that using powerful LLMs alone is not a silver bullet solution, as they may still exhibit flawed reasoning despite producing correct answers.

\clearpage

\begin{figure*}[]
    \centering
    \begin{tcolorbox}[
        width=0.95\textwidth,
        colframe=black!60,
        colback=gray!5,
        fonttitle=\bfseries,
        fontupper=\footnotesize,
        title={Model Response Prompt},
        breakable,
        enhanced,
        sharp corners=southwest,
        boxrule=0.5pt,
        left=2mm, right=2mm, top=1mm, bottom=1mm
    ]    
    
    Please solve the following math problem step by step, and provide a clear final answer in following format:\\ 
    The final answer is \texttt{\textbackslash boxed\{\}}.\\
    
    \noindent
    \textbf{[Problem]} \\
    \{problem\}\\
    
    \noindent
    \textbf{[Solution]} \\
    Let's think step by step:   
    \medskip

    \end{tcolorbox}
    \vspace{-0.2cm}
    \caption{Model response prompt.}
    \label{fig:response_prompt}
\end{figure*}

\begin{figure*}[]
    \centering
    \begin{tcolorbox}[
        width=0.95\textwidth,
        colframe=black!60,
        colback=gray!5,
        fonttitle=\bfseries,
        fontupper=\footnotesize,
        title={Elite Merging},
        breakable,
        enhanced,
        sharp corners=southwest,
        boxrule=0.5pt,
        left=2mm, right=2mm, top=1mm, bottom=1mm
    ]    
    
    \textbf{Instruction}:

You are a mathematics reasoning critique and fusion expert. Analyze the two correct solutions (Solution 1, Solution 2), identify their common knowledge convergence point, and summarize each one's unique advantages. Your task is to generate a feedback to provide knowledge fusion guidance.\\

\textbf{Case for Analysis}:\\
    \noindent
    \textbf{[Math Question]}\\
    \{problem\}\\

    \noindent
    \textbf{[Math Solution 1 (S1)]}\\
    \{solution\_1\}\\

    \noindent
    \textbf{[Math Solution 2 (S2)]}\\
    \{solution\_2\}\\

\textbf{Task Details}:\\
1. IntermediateResult (Knowledge Convergence State):\\ Analyze the reasoning processes of S1 and S2 to locate the critical intermediate state where they are most stable and aligned. Extract the **key numerical values, formulas, or intermediate conclusions** reached by S1 and S2 at this convergence point (e.g., "h=5" or "Eigenvalues of Matrix A are {{1, 2}}").\\
2. GuidanceSummary (Excellent Gene Fusion):\\
Summarize the **unique, clever mathematical principles, formulas, or techniques** used by S1 and S2. Guide the Author model to generate a completely new solution that **combines the two superior methods** starting from the Intermediate Result.\\

\textbf{Required Output Format}:\\
  "IntermediateResult": "Provide the key intermediate result extracted from the convergence point, e.g., 'The intermediate conclusion from S1 is x=5, and from S2 is y=10.'",\\
  "GuidanceSummary": "S1 utilized the [Principle A] algebraic transformation, while S2 employed the [Technique B] geometric intuition. Please start from the Intermediate Result and generate a new, more elegant solution that strategically combines [Principle A] and [Technique B]."

    \end{tcolorbox}
    \vspace{-0.2cm}
    \caption{Elite merging for feedback generation during crossover.}
    \label{fig:corss_cri_prompt}
\end{figure*}

\begin{figure*}[]
    \centering
    \begin{tcolorbox}[
        width=0.95\textwidth,
        colframe=black!60,
        colback=gray!5,
        fonttitle=\bfseries,
        fontupper=\footnotesize,
        title={Success-Error Fusion},
        breakable,
        enhanced,
        sharp corners=southwest,
        boxrule=0.5pt,
        left=2mm, right=2mm, top=1mm, bottom=1mm
    ]    
    
    \textbf{Instruction:}\\
    You are a mathematics reasoning critique and correction expert. Analyze one correct solution (Correct Solution) and one wrong solution (Wrong Solution). Your task is to diagnose the flaw in the wrong solution and extract the key logic and knowledge state of the correct solution, generating a feedback for guidance.\\

    \textbf{Case for Analysis}:\\
    \noindent
    \textbf{[Math Question]}\\
    \{problem\}\\

    \noindent
    \textbf{[Correct Solution (SC)]}\\
    \{solution\_1\}\\

    \noindent
    \textbf{[Wrong Solution (SW)]}\\
    \{solution\_2\}\\
        \end{tcolorbox}
\end{figure*}

\begin{figure*}[]
    \centering
    \begin{tcolorbox}[
        width=0.95\textwidth,
        colframe=black!60,
        colback=gray!5,
        fonttitle=\bfseries,
        fontupper=\footnotesize,
        title={Success-Error Fusion},
        breakable,
        enhanced,
        sharp corners=southwest,
        boxrule=0.5pt,
        left=2mm, right=2mm, top=1mm, bottom=1mm
    ]    

\textbf{Task Details}:\\
1.  IntermediateResult (Knowledge Convergence State):\\
Extract the **key intermediate result** reached by SC at the **most stable/accurate** convergence point. This result must serve as the foundational knowledge state for the new solution.\\
2. GuidanceSummary (Error Diagnosis and Path Correction):\\
Diagnose the **specific error step** in SW (e.g., calculation error, formula misuse, logical flaw) and summarize the **core correct formula or logic** from SC. Guide the Author model to **avoid** SW's error and **adopt** SC's correct logic, continuing from the Intermediate Result.\\

\textbf{Required Output Format}:\\
"IntermediateResult": "Extract the key intermediate result from the Correct Solution at the knowledge convergence point, e.g., 'The area to be computed is 1/2*base*height.'",\\
"GuidanceSummary": "The Wrong Solution made an error in [Step X] by misusing [Formula A]. Please continue from the Intermediate Result, strictly avoiding this error, and adopt the [Key Formula B] used by the Correct Solution to complete the correct reasoning."
        \end{tcolorbox}
        \vspace{-0.2cm}
        \caption{Success-error fusion for feedback generation during crossover.}
\end{figure*}

\begin{figure*}[]
    \centering
    \begin{tcolorbox}[
        width=0.95\textwidth,
        colframe=black!60,
        colback=gray!5,
        fonttitle=\bfseries,
        fontupper=\footnotesize,
        title={Failure Pattern Summary},
        breakable,
        enhanced,
        sharp corners=southwest,
        boxrule=0.5pt,
        left=2mm, right=2mm, top=1mm, bottom=1mm
    ]

\textbf{Instruction}:\\
You are a mathematics reasoning critique and exploration expert. Analyze the two wrong solutions (Solution 1, Solution 2). Your task is to diagnose their distinct fundamental errors, extract their consensus intermediate result, and generate a feedback to guide the Author model toward exploration.\\

\textbf{Case for Analysis}:\\
    \noindent
    \textbf{[Math Question]}\\
    \{problem\}\\

    \noindent
    \textbf{[Math Solution 1 (S1)]}\\
    \{solution\_1\}\\

    \noindent
    \textbf{[Math Solution 2 (S2)]}\\
    \{solution\_2\}\\

\textbf{Task Details}:\\
1.IntermediateResult (Knowledge Convergence State):\\
Extract the **key intermediate result** reached by S1 and S2 at the convergence point (where total entropy is lowest). This result represents their shared, most stable knowledge state.\\
2.GuidanceSummary (Error Avoidance and New Path Exploration):\\
Diagnose the **distinct fundamental cause of error** for both S1 and S2 (e.g., S1 has a computation error, S2 has a formula error). Generate the feedback to **avoid both known errors** and to **explore a totally new** reasoning path from the Intermediate Result.\\

\textbf{Required Output Format}:\\
"IntermediateResult": "Extract the intermediate result reached by S1 and S2 at the consensus convergence point, e.g., 'Two equations have been derived: 2x+3y=7 and x-y=1.'",\\
"GuidanceSummary": "S1 incorrectly performed [Algebraic Calculation A], and S2 incorrectly applied [Formula B]. Please continue from the Intermediate Result, strictly avoiding both of these known errors, and explore a unique reasoning path using a [Different Mathematical Method/Technique]."

    \end{tcolorbox}
    \vspace{-0.2cm}
    \caption{Failure pattern summary for feedback generation during crossover.}
    \label{fig:cross_refine_prompt}
\end{figure*}

\begin{figure*}[]
    \centering
    \begin{tcolorbox}[
        width=0.95\textwidth,
        colframe=black!60,
        colback=gray!5,
        fonttitle=\bfseries,
        fontupper=\footnotesize,
        title={Crossover Prompt},
        breakable,
        enhanced,
        sharp corners=southwest,
        boxrule=0.5pt,
        left=2mm, right=2mm, top=1mm, bottom=1mm
    ]    

\textbf{Instruction}:\\
You are the author in a collaborative process to improve a mathematical solution. Your task is to take into account the two candidate solutions [Math Solution 1/2] and the feedback from the critic [Criticism and Feedback] to create an improved, refined solution. Follow these steps:\\

- Review the critic’s feedback: Carefully read the critique provided by the critic. Understand the key points of improvement, including errors, missing steps, or areas where clarity can be enhanced. \\

- Incorporate the feedback: Modify the candidate solutions based on the feedback. This may involve correcting mistakes, simplifying or clarifying steps, or adopting a more efficient approach where suggested by the critic.

    \end{tcolorbox}
    \vspace{-0.2cm}
    \label{fig:mutation_cri_prompt}
\end{figure*}

\begin{figure*}[]
    \centering
    \begin{tcolorbox}[
        width=0.95\textwidth,
        colframe=black!60,
        colback=gray!5,
        fonttitle=\bfseries,
        fontupper=\footnotesize,
        title={Crossover Prompt},
        breakable,
        enhanced,
        sharp corners=southwest,
        boxrule=0.5pt,
        left=2mm, right=2mm, top=1mm, bottom=1mm
    ]    

- Create a final, refined solution: Using the feedback and the insights from the critique, combine the best aspects of the candidate solutions to generate a final solution. Ensure that your solution is mathematically sound, clear, and concise.\\

Your solution should be well-structured, easy to parse (the answer is in \\boxed), and fully address any issues pointed out by the critic. It should reflect improved accuracy, efficiency, and clarity based on the previous feedback.\\

\textbf{Case to be improved}:\\
Please provide an appropriate [Refined Solution] for the following case.\\

\noindent
    \textbf{[Math Question]}\\
    \{problem\}\\

    \noindent
    \textbf{[Math Solution 1]}\\
    \{solution\_1\}\\

    \noindent
    \textbf{[Math Solution 2]}\\
    \{solution\_2\}\\

    \noindent
    \textbf{[Criticism and Feedback]}\\
    \{critic\_feedback\}\\

\textbf{Response}:\\
Please strictly follow the structured process below and output your response in the format:\\
step1:\\
step2:\\
step3:\\
...\\
(Use at most 10 steps in total.)

    \end{tcolorbox}
    \vspace{-0.2cm}
    \caption{Crossover prompt for generating offsprings.}
    \label{fig:mutation_refine_prompt}
\end{figure*}

\begin{figure*}[]
    \centering
    \begin{tcolorbox}[
        width=0.95\textwidth,
        colframe=black!60,
        colback=gray!5,
        fonttitle=\bfseries,
        fontupper=\footnotesize,
        title={Uncertainty-guided Mutation Prompt (Local)},
        breakable,
        enhanced,
        sharp corners=southwest,
        boxrule=0.5pt,
        left=2mm, right=2mm, top=1mm, bottom=1mm
    ]    
\textbf{Instruction}:\\
You are a mathematics expert completing a problem-solving process. You have analyzed your previous attempts and identified the last step you completed as a reliable starting point. Your task is to continue the mathematical derivation from this point to arrive at the correct final answer.\\

Carefully analyze the given problem and the provided partial solution. Then, **continue the solution** by constructing the remaining steps.\\

- **DO NOT** restart the problem from the beginning.\\
- **DO NOT** repeat the steps provided in the partial solution.\\
- The derived path must be mathematically rigorous and logically sound, leading directly to the correct final answer.\\
- Structure your remaining reasoning clearly, continuing the step-by-step articulation.\\

\textbf{Case to be completed}:\\
    \noindent
    \textbf{[Math Question]}\\
    \{problem\}\\

    \noindent
    \textbf{[Partial Solution Prefix]}\\
    \{partial\_solution\_prefix\}\\

    \noindent
    \textbf{[Math Answer]}\\
    \{answer\}\\

\textbf{Response}:

    \end{tcolorbox}
    \vspace{-0.2cm}
    \caption{Uncertainty-guided mutation prompt (Local) for generating offsprings.}
    \label{fig:mutation_rewrite}
\end{figure*}

\begin{figure*}[]
    \centering
    \begin{tcolorbox}[
        width=0.95\textwidth,
        colframe=black!60,
        colback=gray!5,
        fonttitle=\bfseries,
        fontupper=\footnotesize,
        title={Uncertainty-guided Mutation Prompt (Global)},
        breakable,
        enhanced,
        sharp corners=southwest,
        boxrule=0.5pt,
        left=2mm, right=2mm, top=1mm, bottom=1mm
    ]    
\textbf{Instruction}:\\
You are a mathematics expert. A previous attempt to solve the given problem failed at the initial stage, indicating a 

    \end{tcolorbox}
    \label{fig:mutation_rewrite}
\end{figure*}

\begin{figure*}[]
    \centering
    \begin{tcolorbox}[
        width=0.95\textwidth,
        colframe=black!60,
        colback=gray!5,
        fonttitle=\bfseries,
        fontupper=\footnotesize,
        title={Uncertainty-guided Mutation Prompt (Global)},
        breakable,
        enhanced,
        sharp corners=southwest,
        boxrule=0.5pt,
        left=2mm, right=2mm, top=1mm, bottom=1mm
    ]    
fundamental flaw in the chosen approach. Your task is to generate a **completely new and distinct** solution for the math problem that leads to the provided reference answer.\\

- Your solution must arrive at the correct final answer provided.\\
- Your approach must be **distinct** from any known previous attempts while remaining mathematically rigorous and logically sound.\\
- The reasoning should be well-structured and articulated step by step.\\

\textbf{Case to be solved}:\\

    \noindent
    \textbf{[Math Question]}\\
    \{problem\}\\

    \noindent
    \textbf{[Correct Answer]}\\
    \{correct\_answer\}\\

\textbf{Response}:

    \end{tcolorbox}
    \vspace{-0.2cm}
    \caption{Uncertainty-guided mutation prompt (Global) for generating offsprings.}
    \label{fig:mutation_rewrite}
\end{figure*}

\begin{figure*}[]
    \centering
    \begin{tcolorbox}[
        width=0.95\textwidth,
        colframe=black!60,
        colback=gray!5,
        fonttitle=\bfseries,
        fontupper=\footnotesize,
        title={GPT4.1 Judgement Prompt},
        breakable,
        enhanced,
        sharp corners=southwest,
        boxrule=0.5pt,
        left=2mm, right=2mm, top=1mm, bottom=1mm
    ]    

     \textbf{Instruction:}\\
     Act as a professional math evaluator. You will compare two different student solutions to the same math question and determine which one is better.\\

    \noindent
     \textbf{[Question]}\\
     \{question\}\\

    \noindent
     \textbf{[Solution A]}\\
     \{solution\_a\}\\

    \noindent
     \textbf{[Solution B]}\\
     \{solution\_b\}\\

     Your evaluation consists of the following steps:\\

     1. **Correctness Check**: For each solution, state whether the final answer is correct or not, and whether there are any logical/math errors.\\
     2. **Step Quality Comparison**: Evaluate the clarity, logical flow, mathematical soundness, and efficiency of the steps in each solution.\\
     3. **Overall Comparison**: Decide which solution is better based on correctness, clarity, and reasoning quality.\\
     4. **Preference Reasoning**: Clearly consider the main reasons why one solution is preferred over the other.\\

     Please Note: Do not explain your reasoning.\\
     Only output in the following format:\\

     Better Solution: A\\
     or\\
    Better Solution: B\\
     or\\
     Better Solution: Tie\\

     \textbf{Response:}

    \medskip

    \end{tcolorbox}
    \vspace{-0.2cm}
    \caption{GPT4.1 judge prompt.}
    \label{fig:gpt4.1_prompt}
\end{figure*}

\begin{figure*}[]
    \centering
    \begin{tcolorbox}[
        width=0.95\textwidth,
        colframe=black!60,
        colback=gray!5,
        fonttitle=\bfseries,
        fontupper=\footnotesize,
        title={Case Study: Our Evolutionary Solution vs Distilled Solution from Deepseek R1},
        breakable,
        enhanced,
        sharp corners=southwest,
        boxrule=0.5pt,
        left=2mm, right=2mm, top=1mm, bottom=1mm
    ]

    \textbf{Math Question:} A student steps onto a stationary elevator and stands on a bathroom scale. The elevator then travels from the top of the building to the bottom. The student records the reading on the scale as a function of time.

    How tall is the building? $\textbf{(A) } 50 \text{ m} \quad
    \textbf{(B) } 80 \text{ m} \quad
    \textbf{(C) } 100 \text{ m} \quad
    \textbf{(D) } 150 \text{ m} \quad
    \textbf{(E) } 400 \text{ m}$ \\

    \end{tcolorbox}

\end{figure*}

\begin{figure*}[]
    \centering
    \begin{tcolorbox}[
        width=0.95\textwidth,
        colframe=black!60,
        colback=gray!5,
        fonttitle=\bfseries,
        fontupper=\footnotesize,
        title={Case Study: Our Evolutionary Solution vs Distilled Solution from Deepseek R1},
        breakable,
        enhanced,
        sharp corners=southwest,
        boxrule=0.5pt,
        left=2mm, right=2mm, top=1mm, bottom=1mm
    ]

        \textbf{Our Evolutionary Solution: } \\
    To determine the height of the building, we need to analyze the motion of the student in the elevator. We will use the information given about the scale reading to find the acceleration of the elevator and then use that to calculate the height of the building.\\

    1. **Identify the forces acting on the student:**\\
       - The gravitational force acting downward is \( mg \).\\
       - The normal force \( N \) from the scale acting upward.\\
    
    2. **Set up the equations of motion:**\\
       - When the elevator is stationary or moving at a constant speed, the scale reading is equal to the student's weight, \( N = mg \).\\
       - When the elevator is accelerating, the scale reading changes. If the elevator is accelerating downward, the scale reading will be less than \( mg \). If the elevator is accelerating upward, the scale reading will be greater than \( mg \).\\
    
    3. **Analyze the given data:**\\
       - The student records the scale reading as a function of time. Let's assume the scale reading at \( t = 0 \) seconds is \( mg \) (when the elevator is stationary).\\
       - At some point, the scale reading increases to \( 1.2mg \). This indicates that the elevator is accelerating upward.\\
       - At some point, the scale reading decreases to \( 0.8mg \). This indicates that the elevator is decelerating (accelerating downward).\\
    
    4. **Determine the acceleration of the elevator:**\\
       - When the scale reading is \( 1.2mg \), the net force on the student is \( 1.2mg - mg = 0.2mg \). According to Newton's second law, \( F = ma \), so \( 0.2mg = ma \). Solving for \( a \), we get \( a = 0.2g \).\\
       - When the scale reading is \( 0.8mg \), the net force on the student is \( 0.8mg - mg = -0.2mg \). According to Newton's second law, \( F = ma \), so \( -0.2mg = ma \). Solving for \( a \), we get \( a = -0.2g \).\\
    
    5. **Calculate the height of the building:**\\
       - The elevator starts from rest and accelerates upward at \( 0.2g \) for some time \( t_1 \), then decelerates at \( 0.2g \) for the same time \( t_1 \) to come to rest at the bottom.\\
       - The total time for the trip is \( T = 2t_1 \).\\
       - The distance traveled during the acceleration phase is \( d_1 = \frac{1}{2} a t_1^2 = \frac{1}{2} (0.2g) t_1^2 = 0.1g t_1^2 \).\\
       - The distance traveled during the deceleration phase is the same, \( d_2 = 0.1g t_1^2 \).\\
       - The total distance traveled is \( d = d_1 + d_2 = 0.1g t_1^2 + 0.1g t_1^2 = 0.2g t_1^2 \).\\
    
    6. **Determine \( t_1 \):**\\
       - The maximum speed of the elevator occurs at the end of the acceleration phase, which is \( v = a t_1 = 0.2g t_1 \).\\
       - The time \( t_1 \) can be found by considering the total time \( T = 2t_1 \). Since the elevator starts and ends at rest, the total time is the time it takes to reach the maximum speed and then return to rest. Therefore, \( t_1 = \frac{T}{2} \).\\
    
    7. **Calculate the height of the building:**\\
       - The height of the building is \( h = d = 0.2g t_1^2 \).\\
       - Since the total time \( T \) is not given directly, we need to use the information that the scale reading changes at specific times. However, we can use the fact that the maximum speed is the same for both the acceleration and deceleration phases.\\
       - The height of the building can also be found by considering the total distance traveled in terms of the maximum speed. The maximum speed is \( v = 0.2g t_1 \), and the distance traveled is \( h = v t_1 = 0.2g t_1^2 \).\\
    
    8. **Given the correct answer is 50 meters:**\\
       - We need to find \( t_1 \) such that \( h = 50 \) meters.\\
       - Therefore, \( 50 = 0.2g t_1^2 \).
       - Solving for \( t_1 \), we get \( t_1^2 = \frac{50}{0.2g} = \frac{50}{0.2 \times 9.8} = \frac{50}{1.96} \approx 25.51 \).
       - \( t_1 \approx \sqrt{25.51} \approx 5.05 \) seconds.\\
    
    9. **Verify the height:**\\
       - The height of the building is \( h = 2 \times \frac{1}{2} (0.2g) t_1^2 = 0.2g t_1^2 = 0.2 \times 9.8 \times 25.51 \approx 50 \) meters.\\
    
    Therefore, the height of the building is \(\boxed{50}\) meters.\\

    \end{tcolorbox}
    \vspace{-0.2cm}
\label{fig:case_study}
\end{figure*}

\clearpage

\begin{figure*}[t] 
\centering
\begin{tcolorbox}[
    width=0.98\textwidth, 
    colframe=black!60,
    colback=gray!5,
    fonttitle=\bfseries,
    fontupper=\footnotesize,
    title={Case Study: Our Evolutionary Solution vs Distilled Solution from Deepseek R1},
    enhanced,
    sharp corners=southwest,
    boxrule=0.5pt,
    left=2mm, right=2mm, top=1mm, bottom=1mm
]

\textbf{Distilled Solution from Deepseek R1:}\\
Let the mass of the student be $m$. The weight of the student is $mg = 700\,\text{N}$, so $m = 700/9.8 \approx 71.4\,\text{kg}$.\\

The reading on the scale is $N = m(g + a)$, where $a$ is the elevator’s upward acceleration.\\

\textit{Phase 1 (0--2 s):} Scale reading $595\,\text{N}$.\\
$595 = 71.4(9.8 + a_1) \Rightarrow a_1 = -1.467\,\text{m/s}^2$ (downward).\\
$d_1 = \tfrac{1}{2}|a_1|t^2 = 2.934\,\text{m}$,
$v_1 = |a_1|t = 2.934\,\text{m/s}$.\\

\textit{Phase 2 (2--12 s):} Scale reading $700\,\text{N}$.\\
Constant velocity $v_2 = 2.934\,\text{m/s}$ for $10\,\text{s}$.\\
$d_2 = v_2 \Delta t = 29.34\,\text{m}$.\\

\textit{Phase 3 (12--15 s):} Scale reading $805\,\text{N}$.\\
$805 = 71.4(9.8 + a_3) \Rightarrow a_3 = 1.474\,\text{m/s}^2$ (upward).\\
Stopping distance:
$0 = (-2.934)^2 + 2(1.474)(-d_3)$,
yielding $d_3 = 2.919\,\text{m}$.\\

Total height: $d_1 + d_2 + d_3 = 35.19\,\text{m}$.\\
This does not match any option, indicating a conceptual or calculation error.\\

\textbf{Final Answer:} $\boxed{50}$

\end{tcolorbox}
\vspace{-0.5em} 
\captionof{figure}{Case study: our evolutionary solution vs distilled solution from Deepseek R1.}
\label{fig:case_study}
\vspace{1em} 
\end{figure*}

\end{document}